\pdfoutput=1

\documentclass[11pt]{article}

\usepackage{emnlp2024}

\usepackage{times}
\usepackage{latexsym}
\usepackage{soul}
\usepackage{lipsum}
\usepackage{graphicx}
\usepackage{xcolor}
\usepackage{stfloats}
\usepackage{lipsum}
\usepackage{amsmath}
\usepackage{animate}
\usepackage{subcaption}
\usepackage{parskip}
\usepackage{algorithm}
\usepackage{algpseudocode}
\usepackage{amsmath}
\usepackage{hyperref}
\usepackage{listings}
\usepackage{float}
\usepackage{xcolor}
\usepackage{placeins}

\definecolor{yifei}{RGB}{255,0,0}

\usepackage[T1]{fontenc}

\usepackage[utf8]{inputenc}

\usepackage{microtype}

\usepackage{inconsolata}
\usepackage{multirow}
\usepackage{booktabs}
\usepackage{multicol}
\usepackage{ragged2e}

\usepackage[export]{adjustbox}

\usepackage{longtable}

\title{SHIELD: LLM-Driven Schema Induction for Predictive Analytics in EV Battery Supply Chain Disruptions}

\author{
Zhi-Qi Cheng\textsuperscript{\rm{1}} \quad 
Yifei Dong\textsuperscript{\rm{2}} \quad 
Aike Shi\textsuperscript{\rm{3}} \quad 
Wei Liu\textsuperscript{\rm{4}} \quad 
Yuzhi Hu\textsuperscript{\rm{5}} \\
\textbf{Jason O'Connor}\textsuperscript{\rm{1}} \quad 
\textbf{Alexander G. Hauptmann}\textsuperscript{\rm{1}} \quad 
\textbf{Kate S. Whitefoot}\textsuperscript{\rm{1}} \vspace{0.5em}\\
\textsuperscript{1}Carnegie Mellon University \
\textsuperscript{2}Columbia University \
\textsuperscript{3}Georgia Institute of Technology \\
\textsuperscript{4}University of Michigan, Ann Arbor \
\textsuperscript{5}Boston University \vspace{0.5em} \\
Project Page: \url{https://f1y1113.github.io/MFI/} 
}

\newcommand\nnfootnote[1]{%
  \begin{NoHyper}
  \renewcommand\thefootnote{}\footnote{#1}%
  \addtocounter{footnote}{-1}%
  \end{NoHyper}
}

\begin{document}
\maketitle

\begin{abstract}
The electric vehicle (EV) battery supply chain's vulnerability to disruptions necessitates advanced predictive analytics.~We present SHIELD (Schema-based Hierarchical Induction for~EV~supply chain Disruption), a system integrating Large Language Models (LLMs) with domain expertise for EV battery supply chain risk assessment.~SHIELD combines:~(1)~LLM-driven schema learning to construct a comprehensive knowledge library,~(2)~a disruption analysis system utilizing fine-tuned language models for event extraction, multi-dimensional similarity matching for schema matching, and Graph Convolutional Networks (GCNs) with logical constraints for prediction, and~(3)~an interactive interface for visualizing results and incorporating expert feedback to enhance decision-making.~Evaluated on 12,070 paragraphs from 365 sources (2022-2023), SHIELD outperforms baseline GCNs and LLM+prompt methods~(e.g.~GPT-4o) in disruption prediction.~These results demonstrate SHIELD's effectiveness in combining LLM capabilities with domain expertise for enhanced supply chain risk assessment.
\end{abstract}


\nnfootnote{Completed by Y. Dong and Y. Hu during remote visits, and A. Shi and W. Liu during CMU internships. Z. Cheng, Y. Dong, A. Shi, W. Liu, and Y. Hu contributed equally. J. O'Connor, A. Hauptmann, and K. Whitefoot provided guidance. See Appx.~\ref{app:contribution} for details. Correspondence: \href{mailto@cs.cmu.edu,alex@cs.cmu.edu}{zhiqic,alex@cs.cmu.edu}, \href{mailto@andrew.cmu.edu}{kwhitefoot@andrew.cmu.edu}.}

\vspace{-1cm}

\section{Introduction}

The expected widespread adoption of electric vehicles (EVs) is threatened by risks associated with the geographic and economic concentration of critical battery minerals, such as lithium, cobalt, and nickel. To enhance the resilience of the EV battery supply chain, manufacturers must anticipate disruptions caused by natural disasters and geopolitical tensions. Proactive strategies and supply diversification are essential to mitigate these risks\footnote{\url{https://nncta.org/_files/documents/chapter4-energy-critical-materials.pdf}}.

\begin{figure}[!t]
    \centering
    \includegraphics[width=0.8\linewidth]{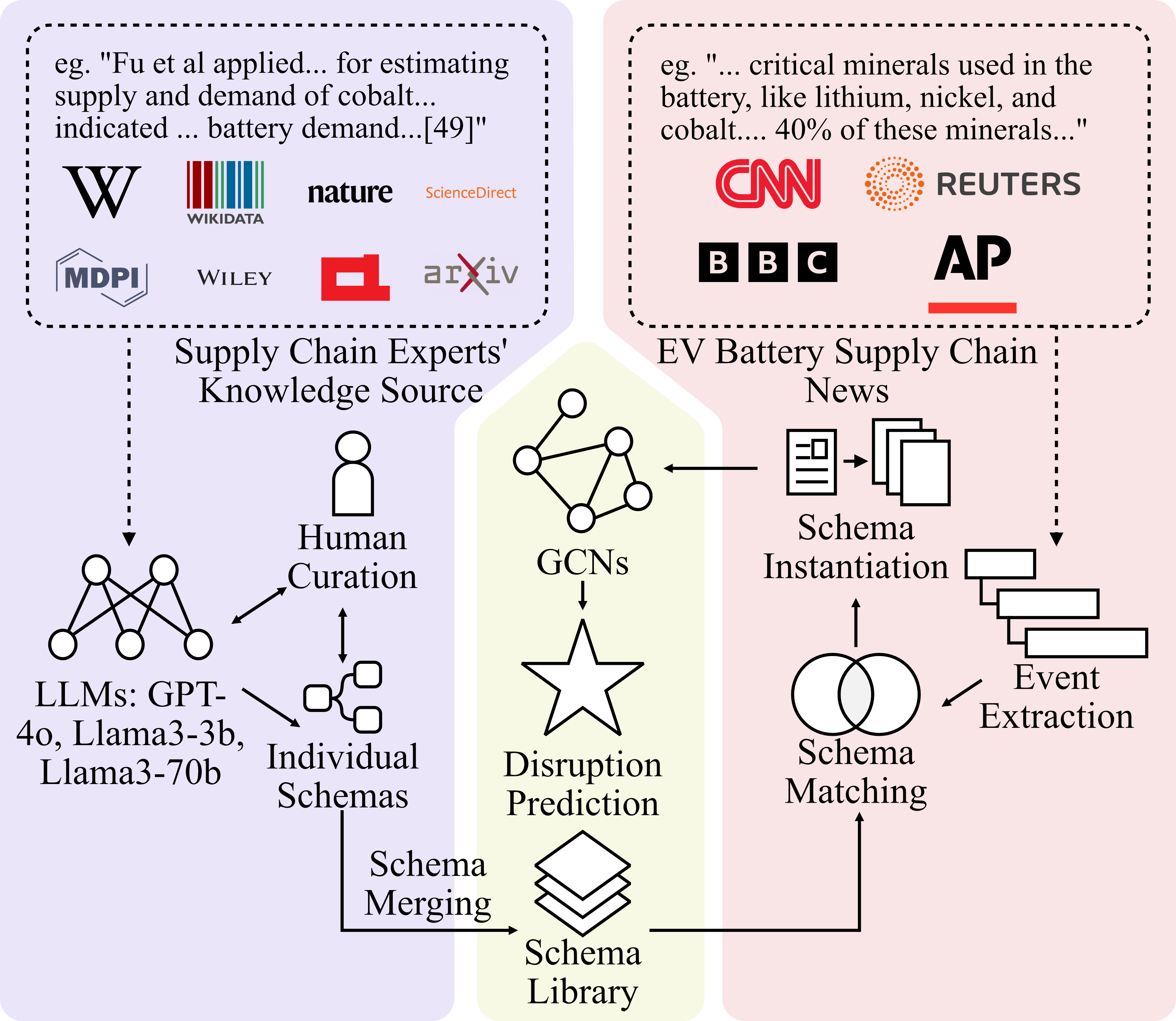}
    \vspace{-2mm}
    \caption{SHIELD's process for EV battery supply chain disruption prediction. The framework integrates LLM-driven schema learning with expert curation, enabling robust event extraction and prediction from diverse news sources.~This approach uniquely combines LLM capabilities with domain expertise, enhancing both predictive accuracy and interpretability for proactive supply chain risk management.}
    \vspace{-6mm}
    \label{fig:motivation}
\end{figure}


Traditional supply chain risk management approaches, which rely on rule-based reasoning and agent-based simulations~\cite{gallab2019risk, pino2010supply, giannakis2011multi, giannakis2016multi, blos2015application}, often fall short in predictive accuracy and adaptability to dynamic market conditions. While machine learning (ML) and deep learning (DL) techniques have enhanced predictive performance~\cite{hegde2020applications, ruz2020sentiment, aljohani2023predictive, silva2017improving, garvey2015analytical, carbonneau2008application}, they frequently sacrifice interpretability, limiting their practical application. Recent studies employing large language models (LLMs) in supply chain management~\cite{ray2023leveraging, wang2022schema, du2022resin, shi2024language, dror2022zero, li2023open} have focused on improving predictions but struggle to fully grasp complex domain-specific supply chain knowledge. This limitation often leads to hallucinations and inaccuracies which, coupled with limited interpretability, hinder the generation of actionable insights crucial for effective risk management.

To address these challenges, we introduce SHIELD (Schema-based Hierarchical Induction for EV supply chain Disruption), a two-stage framework that integrates LLMs and domain expertise for predictive analytics in EV battery supply chains (Fig.~\ref{fig:motivation}):
\begin{enumerate}
    \vspace{-1mm}  
    \item \textit{Schema Learning} (Sec.~\ref{sec:schema_learning}): We leverage LLMs (GPT-4o, Llama3-3b, Llama3-70b) to construct a comprehensive schema library—a structured representation of supply chain components and their relationships—from diverse sources. An interactive system integrates expert knowledge, distilling supply chain expertise from specialized documents.~This approach ensures analyses align with domain knowledge, capturing EV battery supply chain complexities for accurate, interpretable predictions that adapt to industry dynamics through continuous refinement.
    \vspace{-1mm}
    \item \textit{Disruption Analysis}~(Sec.~\ref{sec:disruption_analysis}):~Building on our schema learning, we develop a comprehensive disruption prediction system.~This system integrates fine-tuned RoBERTa~\cite{liu2019roberta} for event detection, multi-dimensional similarity for matching, and Graph Convolutional Networks (GCNs) with logical constraints for impact analysis. The resulting end-to-end system enables precise event extraction and reliable predictions in complex supply chains, mitigating LLM hallucination risks while maintaining efficiency.~This approach offers a scalable solution for real-time supply chain risk assessment and mitigation.
    
    \vspace{-1mm}
\end{enumerate}

\begin{figure*}[!ht]
  \centering
    \includegraphics[width=0.9\linewidth]{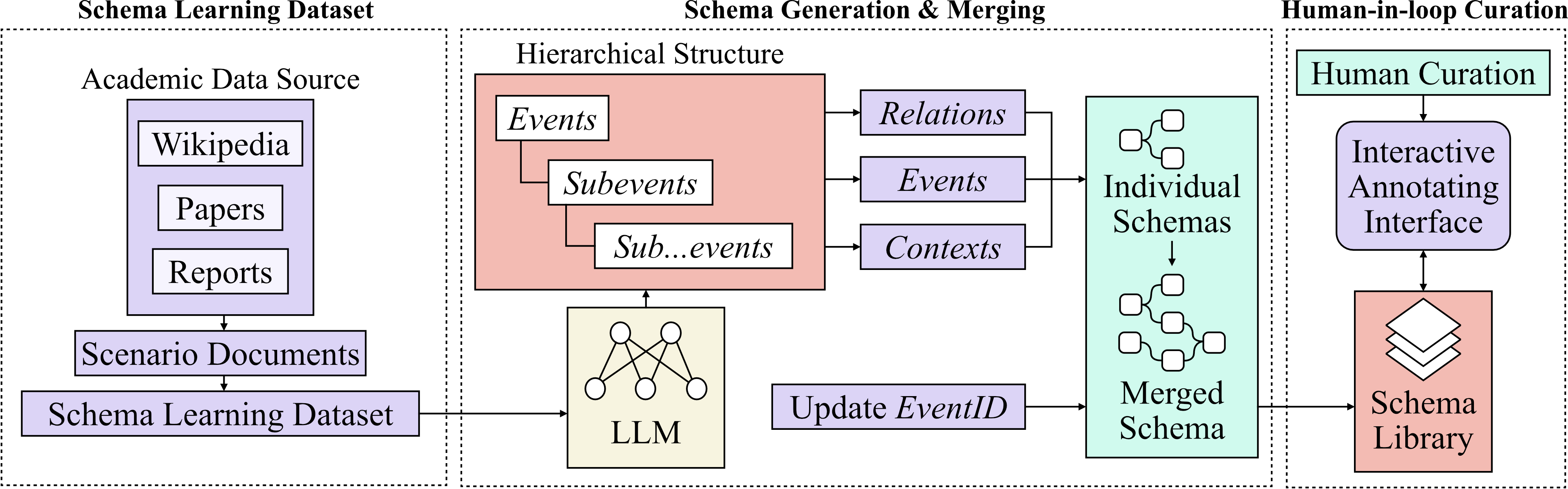}
    \vspace{-2mm}
    \caption{Overview of the supply chain schema construction process, illustrating the collection of diverse sources, schema extraction using large language models, and the integration into a unified schema library.}
    \vspace{-2mm}
    \label{fig:schema_learning}
\end{figure*}

Evaluated on 12,070 paragraphs from 365 sources (2022-2023)~(Sec.~\ref{sec:experiments}), SHIELD outperforms baseline GCNs and LLM+prompt methods~(e.g.,~GPT-4o) in disruption prediction.~By integrating LLM capabilities with domain expertise, this framework enhances supply chain risk assessment.~Key contributions include: (1) an LLM-expert integration methodology for accurate, interpretable predictions; (2) a schema learning and news evaluation dataset spanning the EV battery lifecycle; (3) an interactive schema curation system; and (4) advanced analytical techniques for supply chain analysis.~SHIELD offers a promising approach in supply chain risk management, addressing evolving challenges across the EV industry and beyond.

\section{Related Work}

\noindent \textbf{Supply Chain Risk Management}. AI has been increasingly applied to predict and manage supply chain risks~\cite{ganesh2022future}. Agent-based approaches~\cite{pino2010supply, giannakis2011multi, giannakis2016multi, blos2015application} facilitate inter-agent communication for forecasting but often suffer from limited predictive power and parameter constraints. Rule-based methods~\cite{gallab2019risk, behret2012fuzzy, paul2015supplier, paul2017quantitative, awasthi2018multi, camarillo2018knowledge} offer decision frameworks with minimal quantitative insights. Machine Learning (ML) and Deep Learning (DL) techniques have improved forecasting and disruption prediction~\cite{silva2017improving, hegde2020applications, garvey2015analytical, ruz2020sentiment, aljohani2023predictive, carbonneau2008application}, yet many focus on predictive performance at the expense of interpretability~\cite{hendriksen2023artificial, makridis2023deep}. Recent work has explored large language models (LLMs) in supply chain management~\cite{ray2023leveraging}, but interpretability remains a challenge. Our approach integrates LLMs to enhance both predictive accuracy and interpretability by extracting hierarchical knowledge-graph structures to forecast disruptions.

\noindent \textbf{Schema Induction \& Learning}. Building on early schema induction work~\cite{anderson1979general, evans1967brief}, LLMs~\cite{brown2020language, rae2021scaling} have shown strong schema-learning abilities with minimal supervision. Recent strategies, such as contextual explanations~\cite{wei2021finetuned, lampinen2022can}, rationale-augmented models~\cite{wang2022rationale}, and incremental prompting~\cite{li2023open}, have further refined schema induction. Transformer-based methods~\cite{li2020connecting, li2021future} excel at schema representation through graph structures, with human feedback playing a vital role in improving model accuracy~\cite{mondal2023interactiveie, yang2024human, zhang2023human}. Our method leverages these advancements, combining human feedback with LLM-driven schema induction to improve accuracy and relevance in disruption prediction.

\noindent \textbf{Event Extraction \& Analysis}. Event extraction has evolved from handcrafted features~\cite{ahn2006stages} to neural models like recurrent~\cite{nguyen2016joint, sha2018jointly}, convolutional~\cite{chen2015event}, graph~\cite{zhang2021abstract}, and transformer-based networks~\cite{liu2020event}. Advances in argument extraction~\cite{wang2019hmeae}, zero-shot learning~\cite{huang2018zero}, and weak supervision~\cite{chen2015event} have boosted performance. Our approach enhances event extraction by using fine-tuned RoBERTa models and graph convolutional networks (GCNs) to capture complex event relationships and cascading effects, offering deeper insights into supply chain disruptions compared to traditional methods.

\section{Schema Learning for Supply Chain Disruptions}
\label{sec:schema_learning}
\textbf{Schema Learning Dataset}.~Our dataset comprises 239 diverse sources:~200 academic papers, 22 industry reports, and 17 Wikipedia entries (Fig.~\ref{app:academic_dataset_example} and Fig.~\ref{app:academic_dataset_example_wiki}). This collection provides an up-to-date view of the EV battery supply chain, covering advanced battery technologies (e.g. LFP, NiMH), production processes, and six key raw materials.~We categorized events into 8 categories, three with long-term impacts, subdivided into 18 subcategories. Our analysis includes five-year price trends for all materials, correlated with 39 significant supply chain events. Industry expert feedback refined our categorization into 11 main categories with 27 subcategories, each illustrated with 1-2 real-world events (Tab.~\ref{app:event_category}). The academic dataset was distilled from 239 sources to 125 highly relevant entries.~This dataset of over 1,000 events spans the EV battery lifecycle, enabling our methods to acquire expert knowledge for accurate, real-world predictions.~More details are in Appx.~\ref{app:schema_learning_dataset}.

\begin{figure*}[!ht]
  \centering
    \includegraphics[width=0.95\linewidth]{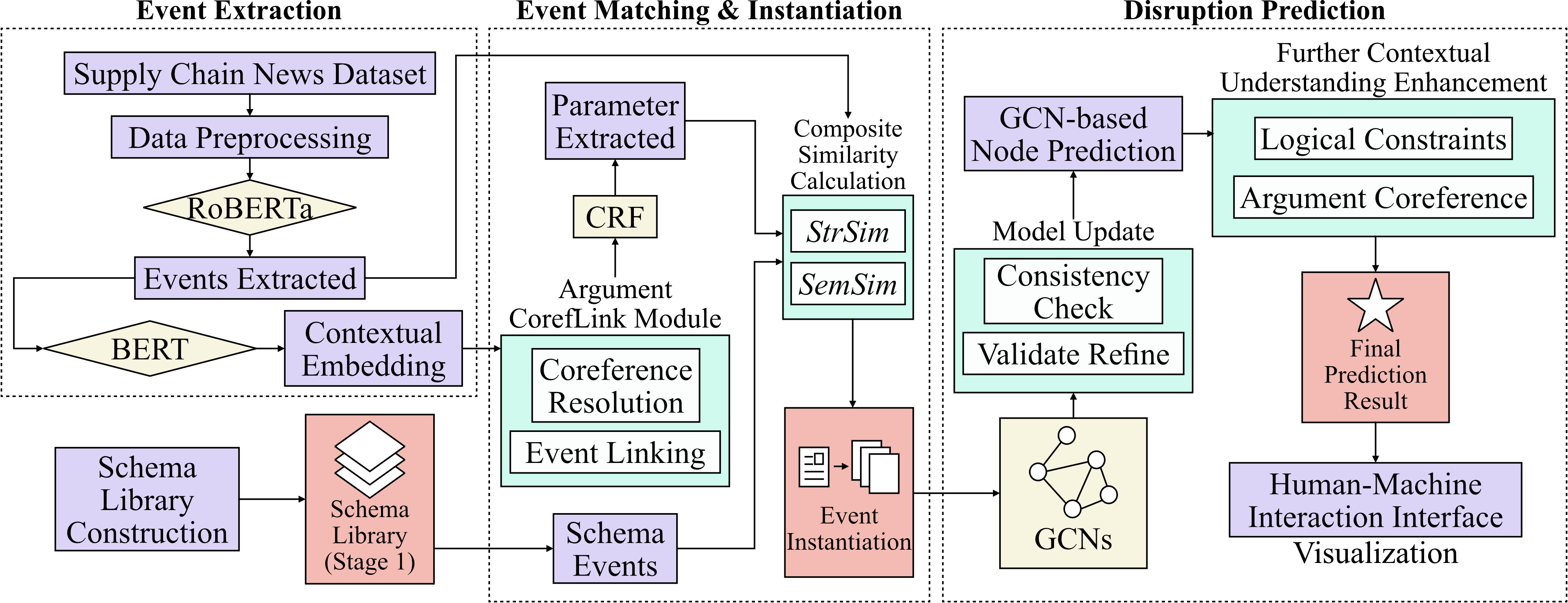}
    \vspace{-2mm}
    \caption{Overview of the supply chain disruption prediction pipeline, illustrating the integration of GCN-based predictions, constrained prediction refinement, and argument coreference resolution.}
    \vspace{-4mm}
    \label{fig:prediction_pipeline}
\end{figure*}

\textbf{Schema Generation \& Merging}. Building upon our collected dataset, our Schema Learning System facilitates the extraction, visualization, and management of schemas from the 125 diverse textual sources (Fig.~\ref{fig:schema_learning}).~The process begins with data cleaning using regular expressions and a locally deployed Llama3-8b model. Subsequently, we employ GPT-4o, Llama3-3b, and Llama3-70b with specific prompts to extract hierarchical structures ($\mathbf{H}$) capturing main events ($\mathbf{E}$) and sub-events ($\mathbf{E}_{sub}$). More details are in Appx.~\ref{app: prompt hierarchical}.

The extracted structures are then converted into individual schemas ($\mathbf{S_i}$) and visualized as graphs, demonstrating the hierarchical nature of the schemas and the relationships between main events and sub-events. These schemas are then integrated into a single library ($\mathbf{S_{final}}$), aggregating contexts ($\mathbf{C_{final}} = \bigcup_{i=1}^{n} \mathbf{C_i}$), merging events ($\mathbf{E_{final}} = \bigcup_{i=1}^{n} \mathbf{E_i}$), and updating event IDs for relations ($\mathbf{R_{final}} = \bigcup_{i=1}^{n} \mathbf{R_i}$). The detailed schema generation and merging algorithm is provided in Appx.~\ref{app: schema merging}.

To ensure efficient retrieval and updates, a dedicated Database \& Storage module manages schema storage, while the Schema Management System incorporates a Schema Viewer, Editor, collaboration tools, and AI-driven suggestions are built to manage and annotate schemas (Appx.~\ref{app: hmi interface}). This human-in-the-loop curated framework streamlines schema extraction and management, enabling interactive knowledge extraction from structured documents, leveraging supply chain experts' insights.

\section{Dynamic Analysis of Supply Chain Disruptions}
\label{sec:disruption_analysis}
\textbf{Supply Chain News Dataset}.~We developed an EV Supply Chain News Dataset (January 2022 - December 2023) to evaluate our system's real-world performance (Appx.~\ref{app:newsdataset}). The dataset comprises 247 articles from major news outlets and 118 enterprise reports from EV battery-related companies (Fig.~\ref{app:newsdataset_example} and Fig.~\ref{app:newsdataset_example_2}). After preprocessing—including text extraction, language standardization, and noise reduction—we obtained Meta data with 3,022 paragraphs. We then fused international news with contemporaneous corporate stories in the meta data, creating 354 diverse documents comprising 12,070 paragraphs.~The final dataset contains approximately 660K words (Table~\ref{app: news_wordcount}), providing a robust foundation for evaluating supply chain disruption detection and analysis.~Comprehensive replication details, including the full dataset and preprocessing pipeline, are provided to ensure reproducibility.

\textbf{Event Extraction}.~Our pipeline extracts multi-faceted events from textual data, focusing on their impact on the EV battery supply chain. We begin with custom-trained SpaCy models\footnote{https://spacy.io/models} for tokenization, sentence segmentation, named entity recognition, and dependency parsing (Appx.~\ref{app: event extraction}).

Building on this, we deploy a fine-tuned RoBERTa model for cross-sentence event detection:
\begin{equation}
\text{EventDetect}_{\text{multi-sentence}}(\mathbf{T}) \rightarrow \mathbf{E_C}
\vspace{-2mm}
\end{equation}
where \(\mathbf{T}\) represents the input text and \(\mathbf{E_C}\) the detected events. These events are then enriched with contextual information using BERT:
\begin{equation}
\text{BERT}_{\text{context}}(\mathbf{E_C}) \rightarrow \mathbf{C_E}
\vspace{-2mm}
\end{equation}
generating contextual embeddings \(\mathbf{C_E}\). To enhance analytical coherence, we implement coreference resolution and event linking:
\begin{equation}
\text{CorefLink}(\mathbf{E_C}) \rightarrow \mathbf{E_L}
\vspace{-2mm}
\end{equation}
This critical step, yielding linked events \(\mathbf{E_L}\), maintains contextual continuity across documents. Subsequently, Conditional Random Fields (CRFs) extract event parameters $\mathbf{P_C}$:
\begin{equation}
\text{CRF}(\mathbf{E_L}) \rightarrow \mathbf{P_C}
\vspace{-2mm}
\end{equation}

Leveraging Graph Convolutional Networks (GCNs), we model complex event relationships and score each event's impact as:
\begin{equation}
\setlength{\abovedisplayskip}{4pt}
\text{ImpactScore}(e_i) = \text{Centrality}(e_i) + \text{Magnitude}(e_i)
\end{equation}
This scoring mechanism balances two crucial factors.~\(\text{Centrality}(e_i)\) represents the event's importance within the network, reflecting its centrality or influence in the supply chain context.~Meanwhile, \(\text{Magnitude}(e_i)\) quantifies the event's impact intensity, indicating its severity or significance.

Finally, we apply logical constraints and argument coreference to ensure robustness:
\begin{equation}
\text{LogicCoref}(\mathbf{P_C}) \rightarrow \mathbf{P_F}
\vspace{-2mm}
\end{equation}
producing a refined, logically consistent set of event parameters \(\mathbf{P_F}\).~More implementation details are in Appx.~\ref{app: event extraction}.

\textbf{Event Matching \& Instantiation}.~We link extracted events with schema library to detect supply chain disruption patterns using a multi-dimensional approach of semantic and structural similarities.~We align each extracted event $E_{\text{ext}} \in \mathbf{E}_{\text{ext}}$ (extracted events) with each schema event $E_{\text{schema}} \in \mathbf{E}_{\textit{schema}}$~(schema events) using a composite similarity:
\begin{equation}
\setlength{\abovedisplayskip}{4pt}
\label{eq:composite_similarity}
\begin{aligned}
    \text{Sim}(E_{\text{ext}}, E_{\text{schema}}) &= \alpha \cdot \text{SemSim}(E_{\text{ext}}, E_{\text{schema}}) \\
    &\quad + \beta \cdot \text{StrSim}(E_{\text{ext}}, E_{\text{schema}})
\end{aligned}
\end{equation}
where \(\text{SemSim}\) captures contextual meaning using BERT embeddings, and \(\text{StrSim}\) assesses structural similarity. Specifically, semantic similarity measures contextual alignment using cosine similarity between BERT embeddings:
\begin{equation}
\label{eq:semantic_similarity}
    \text{SemSim}(E_{\text{ext}}, E_{\text{schema}}) = \frac{\mathbf{v}_{\text{ext}} \cdot \mathbf{v}_{\text{schema}}}{\|\mathbf{v}_{\text{ext}}\| \|\mathbf{v}_{\text{schema}}\|}
\vspace{-2mm}
\end{equation}
where \(\mathbf{v}_{\text{ext}}\) and \(\mathbf{v}_{\text{schema}}\) are BERT embeddings of extracted and schema events. Similarly, structural similarity evaluates parameter overlap using Jaccard similarity:
\begin{equation}
\label{eq:structural_similarity}
    \text{StrSim}(E_{\text{ext}}, E_{\text{schema}}) = \frac{|\mathbf{P}_{\text{ext}} \cap \mathbf{P}_{\text{schema}}|}{|\mathbf{P}_{\text{ext}} \cup \mathbf{P}_{\text{schema}}|}
\vspace{-2mm}
\end{equation}
where \(\mathbf{P}_{\text{ext}}\) and \(\mathbf{P}_{\text{schema}}\) are the parameter sets for the extracted and schema events.

Following the calculation of semantic and structural similarities, we refine matching using heuristic rules from annotated datasets.~Successful matches lead to event instantiation, enriching the event representation with schema attributes:
\begin{equation}
\label{eq:instantiate}
    \text{Instantiate}(E_{\text{matched}}, \mathbf{S}_{\text{schema}}) \rightarrow \mathbf{E}_{\text{inst}}
\vspace{-2mm}
\end{equation}
where \({E}_{\text{matched}}\) represents the matched event, \(\mathbf{S}_{\text{schema}}\) yields the schema library, and \(\mathbf{E}_{\text{inst}}\) refers to the instantiated event with enriched attributes.

To ensure logical adherence to schema constraints, we perform consistency checks. These checks validate the instantiated events against the schema library, ensuring they conform to predefined logical and structural constraints:
\begin{equation}
\label{eq:consistency_check}
    \text{ConsistencyCheck}(\mathbf{E}_{\text{inst}}, \mathbf{S}_{\text{schema}})
\vspace{-2mm}
\end{equation}
This step is crucial for maintaining the integrity of the schema and the reliability of the predictions.

Finally, we incorporate a continuous improvement process through manual review and feedback. Feedback from domain experts is used to update and refine the models, ensuring they adapt to new patterns and maintain high performance.~The complete process is summarized in Algorithm~\ref{alg:event_matching_instantiation}. More implementation details are in Appx.~\ref{app: event matching instantiation}.

\begin{algorithm}[H]
\caption{\small Supply Chain Disruption Prediction}
\label{alg:disruption_prediction}
\small
\begin{algorithmic}[1]
    \State \textbf{Input:} Historical supply chain events $\mathbf{E}$, adjacency matrix $\mathbf{A}$, initial predictions $\hat{y}$
    \State \textbf{Output:} Refined predictions $\hat{y}'$
    
    \State \textbf{GCN-based Prediction} \Comment{Initial prediction using GCN}
    \For{$l=1$ to $L$}
        \State $\mathbf{H}^{(l+1)} = \sigma(\mathbf{A} \mathbf{H}^{(l)} \mathbf{W}^{(l)})$ \Comment{Refer to Eq.~\ref{eq:gcn_propagation}}
    \EndFor
    \State $\hat{y} \leftarrow \mathbf{H}^{(L)}$

    \State \textbf{Constrained Prediction} \Comment{Apply logical constraints}
    \For{each prediction $\hat{y}_i$}
        \State $\hat{y}_i' \leftarrow \text{Constrain}(\hat{y}_i)$
        \State $\quad \text{such that} \quad \mathcal{C}(\hat{y}_i') = \text{true}$ \Comment{Refer to Eq.~\ref{eq:logical_constraints}}
    \EndFor
    
    \State \textbf{Coreference Resolution} \Comment{Link related events}
    \For{each pair of events $(E_i, E_j)$}
        \State $R_{ij} \leftarrow \text{Coref}(E_i, E_j)$ \Comment{Refer to Eq.~\ref{eq:coref_resolution}}
        \If{$R_{ij}$ is coreferential}
            \State Link $E_i$ and $E_j$
        \EndIf
    \EndFor
    \State \textbf{Return:} Refined predictions $\hat{y}$
\end{algorithmic}

\normalsize
\end{algorithm}

\begin{table*}[!ht]
    \centering
    \footnotesize
    \caption{Performance comparison of different LLMs on schema learning in stage 1.}
    \vspace{-2mm}
    \resizebox{0.95\textwidth}{!}{
    \begin{tabular}{lcc|cc|cc}
        \toprule
        & \multicolumn{2}{c|}{ChatGPT4o} & \multicolumn{2}{c|}{Llama3-3b} & \multicolumn{2}{c}{Llama3-70b} \\
        \cmidrule{2-7}
        & Individual Schemas & Integrated Library & Individual Schemas & Integrated Library & Individual Schemas & Integrated Library \\
        \midrule
        Precision & \textbf{0.637} & \textbf{0.184} & 0.198 & 0.018 & 0.353 & 0.019 \\
        Recall    & \textbf{0.695} & \textbf{0.336} & 0.047 & 0.014 & 0.133 & 0.022 \\
        F-score   & \textbf{0.652} & \textbf{0.238} & 0.068 & 0.016 & 0.175 & 0.020 \\
        \bottomrule
    \end{tabular}}
    \label{tab:LLMs_comparison}
\end{table*}

\begin{table}[!ht]
    \centering
    \footnotesize
    \caption{Subjective evaluation by domain experts.}
    \vspace{-2mm}
    \resizebox{0.95\linewidth}{!}{
    \begin{tabular}{lccc}
        \toprule
        Model & Consistency & Accuracy & Completeness \\
        \midrule
        GPT-4o & \textbf{4.5} & \textbf{4.3} & \textbf{4.6} \\
        Llama3-3b  & 1.8 & 1.5 & 1.9 \\
        Llama3-70b & 3.0 & 2.7 & 3.1 \\
        \bottomrule
    \end{tabular}}
    \label{tab:subjective_evaluation}
\end{table}

\textbf{Disruption Prediction}.~Building on the extracted and matched events, we employ Graph Convolutional Networks (GCNs), logical constraints, and argument coreference resolution to predict supply chain disruptions. Note that the events are represented as nodes and interactions as edges using GCNs with the propagation rule:
\begin{equation}
\label{eq:gcn_propagation}
    \mathbf{H}^{(l+1)} = \sigma(\mathbf{A} \mathbf{H}^{(l)} \mathbf{W}^{(l)})
\vspace{-2mm}
\end{equation}
where $\mathbf{H}^{(l)}$ is the hidden state at layer $l$, $\mathbf{A}$ is the adjacency matrix, $\mathbf{W}^{(l)}$ is the weight matrix, and $\sigma$ is a non-linear activation function. We optimize using mean squared error loss with L2 regularization:
\begin{equation}
\label{eq:loss_function}
    \mathcal{L} = \frac{1}{N} \sum_{i=1}^{N} \left( y_i - \hat{y}_i \right)^2 + \lambda \|\mathbf{W}\|^2
\vspace{-2mm}
\end{equation}
where $y_i$ and $\hat{y}_i$ are actual and predicted disruption scores, and $\lambda$ is a regularization parameter. This approach balances prediction accuracy and model complexity, preventing overfitting.

To ensure consistency with domain knowledge, we apply logical constraints, refining initial predictions ($\hat{y}$) to produce final predictions ($\hat{y}'$) that adhere to known rules:
\begin{equation}
\label{eq:logical_constraints}
\begin{aligned}
    \hat{y}' &= \underset{\hat{y}' \in \mathcal{Y}}{\operatorname{arg\,min}} \; \text{Constrain}(\hat{y}) \\
    & \text{subject to} \quad \mathcal{C}(\hat{y}') = \text{true}
\vspace{-2mm}
\end{aligned}
\end{equation}
where $\mathcal{C}$ represents constraint sets. For example, a constraint might ensure that a major supplier's disruption increases risk for dependent manufacturers.

To further enhance the model's contextual understanding, we incorporate argument coreference:
\begin{equation}
\label{eq:coref_resolution}
\begin{aligned}
R_{ij} &= \underset{E_i, E_j \in \mathcal{E}}{\operatorname{arg,max}} ; \text{Coref}(E_i, E_j) \\
& \text{subject to} \quad \text{Coref}(E_i, E_j) = \text{true}
\vspace{-2mm}
\end{aligned}
\end{equation}
where $(E_i, E_j)$ denotes each event pair and $R_{ij}$ represents their relation.
This AllenNLP-based model links entities across event mentions, recognizing when different descriptions refer to the same incident, thereby improving prediction accuracy and context comprehension. Algorithm~\ref{alg:disruption_prediction} outlines our approach, combining GCN-based predictions, logical constraints, and argument coreference resolution. Detailed examples and implementation guidelines are provided in Appx.~\ref{app: disruption prediction}.


\section{Experiments}
\label{sec:experiments}
Our evaluation comprises two parts: (1) \textit{Schema Learning Assessment} and (2) \textit{Supply Chain Disruption Prediction}.~We assess learned schemas against expert knowledge and evaluate our schema induction process's effectiveness in predicting supply chain events. Detailed experimental setup and evaluation metrics are in Appx.~\ref{app: experiment details}.

\subsection{Schema Learning Performance}
\label{exp: schema learning}
We evaluate GPT-4o, Llama3-3b, and Llama3-70b for schema learning, comparing individual schema extraction and integrated library generation.~Tables~\ref{tab:LLMs_comparison} and~\ref{tab:subjective_evaluation} present quantitative metrics and subjective evaluations by domain experts.~GPT-4o outperforms Llama models, achieving F-scores of 0.652 and 0.238 for individual schemas and integrated library generation, respectively. All models perform better in individual schema extraction than integrated library generation, indicating challenges in schema integration.~Subjective assessments align with quantitative metrics, with GPT-4o scoring highest across all criteria (consistency: 4.5, accuracy: 4.3, completeness: 4.6). Individual schemas show strong consistency and completeness but slightly lower accuracy, suggesting a trade-off between comprehensive coverage and precise detail representation.

\subsection{Disruption Detection Performance}
\label{exp: disruption prediction performance}
\textbf{Event Extraction \& Matching}.~Table~\ref{tab:news_quarter_comparison} presents quarterly results for 2022 and 2023 on event extraction and matching using a supply chain news dataset.~Our system maintains consistent performance across quarters, with F-scores ranging from 0.671 to 0.700. This stability suggests robust generalization across different time periods and varying event types. The slight improvement in 2023 (average F-score 0.687 vs 0.683 in 2022) indicates potential refinement in our model's ability to adapt to evolving supply chain dynamics.

\begin{figure*}[!ht]
  \centering
  \begin{subfigure}[b]{0.47\textwidth}
    \centering
    \includegraphics[width=\textwidth]{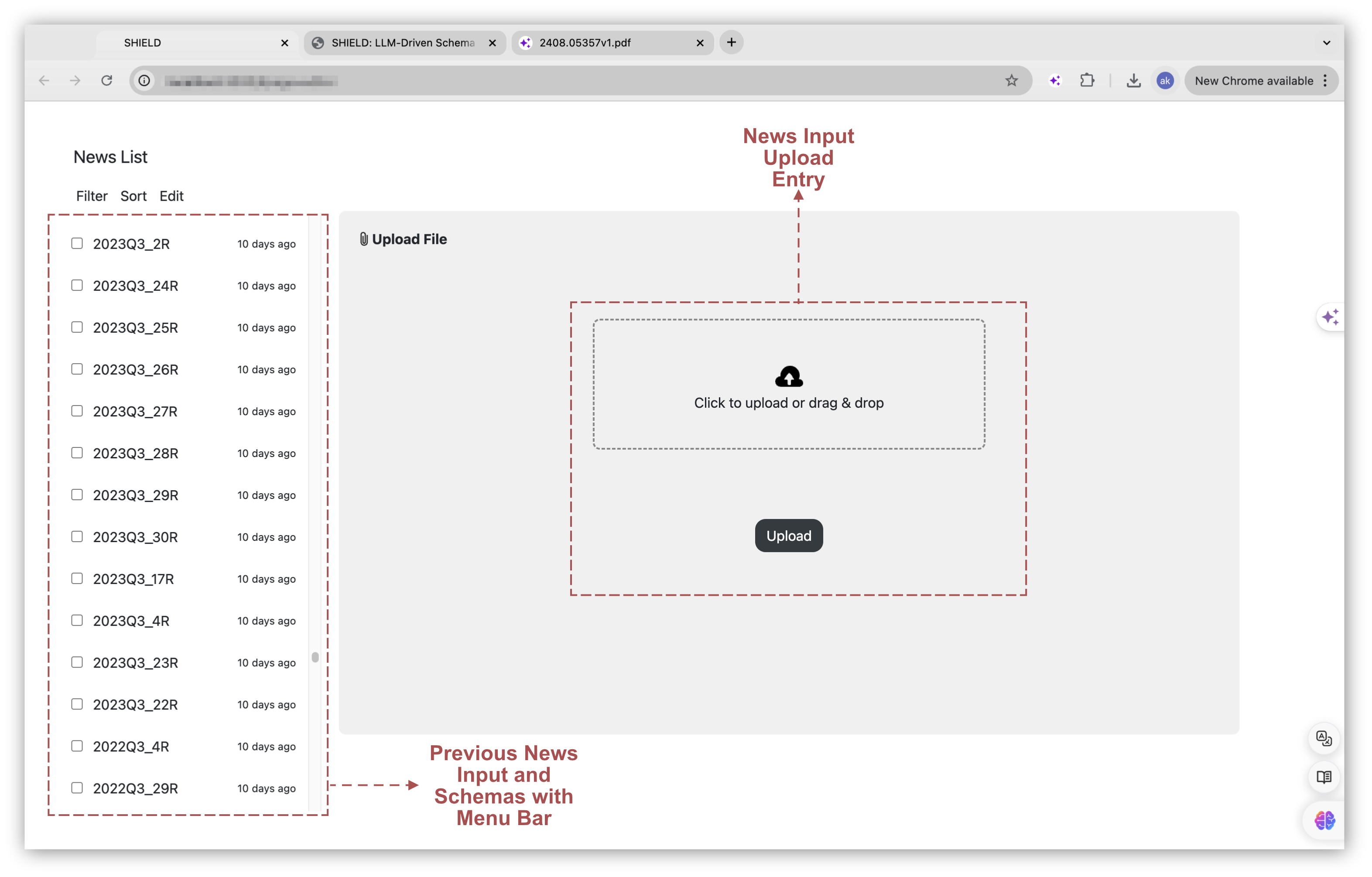}
    \vspace{-4mm}
    \caption{User interface for inputting news reports.}
    \label{fig:user_interface part1 user input}
  \end{subfigure}
  \hfill
  \begin{subfigure}[b]{0.47\textwidth}
    \centering
    \includegraphics[width=\textwidth]{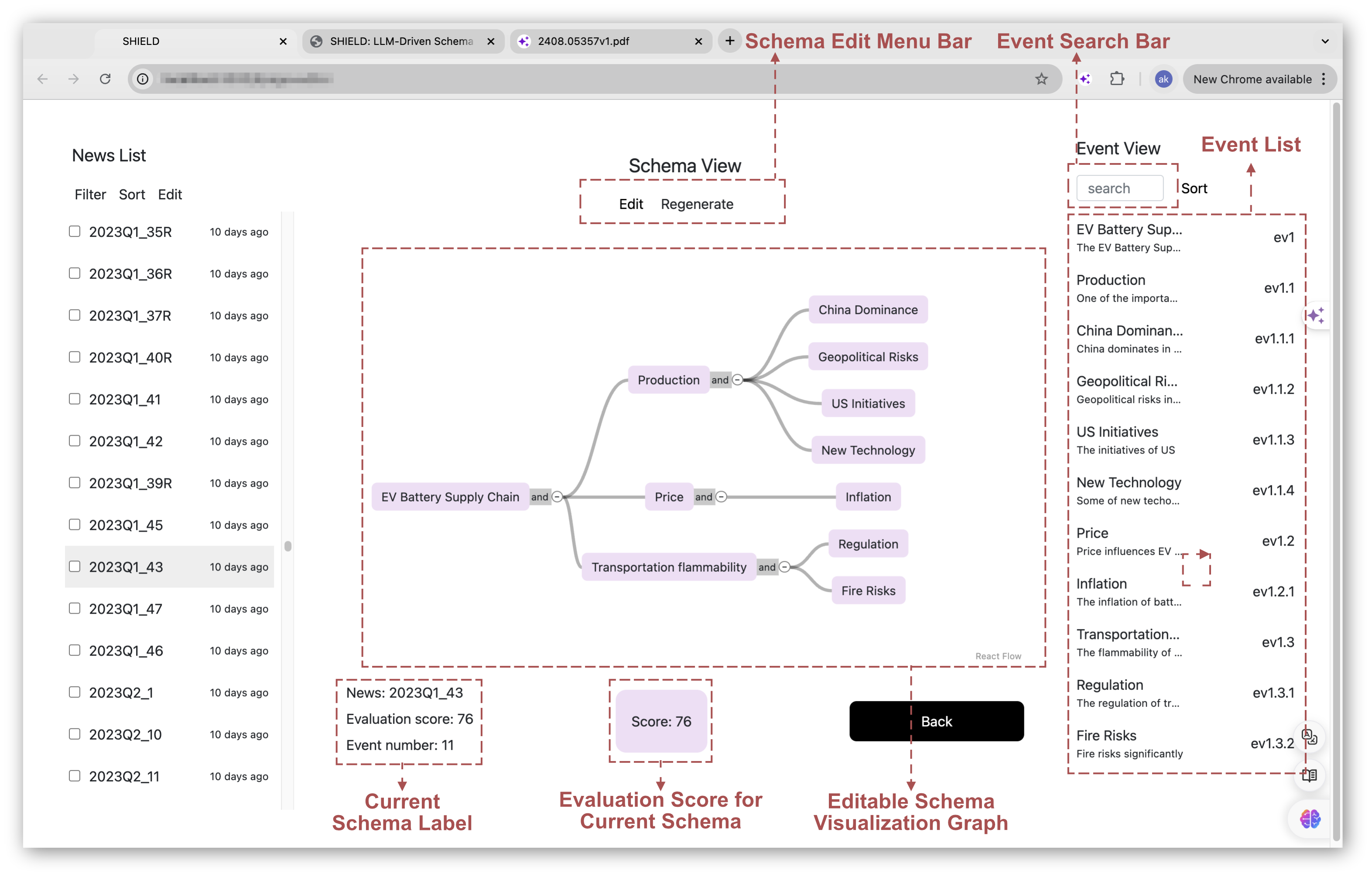}
    \vspace{-4mm}
    \caption{Visualization and editing of final prediction results.}
    \label{fig:user_interface part2 visualization and edit}
  \end{subfigure}
  \vspace{-2mm}
  \caption{User interface for online disruption analysis in stage 2, showing the process from news report input to the visualization and editing of prediction results. More examples are in Appx.~\ref{app: Disruption Prediction User Interface Details}.}
  \vspace{-2mm}
  \label{fig:user_interface}
\end{figure*}

\begin{table*}[!ht]
    \footnotesize
    \centering
    \caption{Event extraction and matching in supply chain news dataset.}
    \vspace{-2mm}
    \resizebox{0.95\textwidth}{!}{
    \begin{tabular}{lcccc|cccc}
        \toprule
        Year & \multicolumn{4}{c|}{2022} & \multicolumn{4}{c}{2023} \\
        \cmidrule{2-9}
        Quarter & Q1 (Jan-Mar) & Q2 (Apr-Jun) & Q3 (Jul-Sep) & Q4 (Oct-Dec) & Q1 (Jan-Mar) & Q2 (Apr-Jun) & Q3 (Jul-Sep) & Q4 (Oct-Dec) \\
        \midrule
        Precision & 0.714 & 0.692 & 0.705 & 0.689 & 0.712 & 0.698 & 0.703 & 0.690 \\
        Recall    & 0.675 & 0.662 & 0.678 & 0.655 & 0.688 & 0.670 & 0.681 & 0.657 \\
        F-score   & 0.694 & 0.677 & 0.691 & 0.671 & 0.700 & 0.684 & 0.692 & 0.673 \\
        \bottomrule
    \end{tabular}}
    \vspace{-2mm}
    \label{tab:news_quarter_comparison}
\end{table*}

\begin{table*}[!ht]
    \centering
    \footnotesize
    \caption{Performance comparison of different models on disruption prediction.}
    \vspace{-2mm}
    \begin{tabular}{lccc}
        \toprule
        Model & Precision & Recall & F-score \\
        \midrule
        Our System (GCNs only) & 0.701 & 0.670 & 0.685 \\
        Our System (GCNs + Logical Constraints) & 0.724 & 0.691 & 0.707 \\
        Our System (GCNs + Logical Constraints + Coreference) & 0.754 & 0.712 & 0.732 \\
        \bottomrule
    \end{tabular}
    \vspace{-2mm}
    \label{tab:disruption_prediction_comparison}
\end{table*}

\begin{table}[!ht]
    \centering
    \footnotesize
    \caption{Performance comparison of direct human interaction with LLMs on disruption prediction.}
    \vspace{-2mm}
    \begin{tabular}{lccc}
        \toprule
        Model & Precision & Recall & F-score \\
        \midrule
        GPT-4o & 0.641 & 0.608 & 0.624 \\
        Llama3-3b & 0.522 & 0.489 & 0.505 \\
        Llama3-70b & 0.557 & 0.523 & 0.540 \\
        Our Method & \textbf{0.754} & \textbf{0.712} & \textbf{0.732} \\
        \bottomrule
    \end{tabular}
    \vspace{-2mm}
    \label{tab:LLM_prompt_comparison}
\end{table}

\textbf{Disruption Detection}.~Our advanced GCNs model, augmented with logical constraints and coreference resolution, was rigorously evaluated against ablation versions and LLM+prompt methods.~Tables~\ref{tab:disruption_prediction_comparison} and~\ref{tab:LLM_prompt_comparison} present the comparative performance metrics.~The full system achieved the highest F-score (0.732), significantly outperforming both ablation versions (GCNs+Logical Constraints:~0.707, GCNs only:~0.685) and LLM+prompt methods (GPT-4o: 0.624).~However, the incremental improvement from the GCNs-only model to our full system (0.685 to 0.732) suggests that while the additional components significantly enhance performance, there remains substantial potential for further optimization in the future.

\subsection{Qualitative Analysis \& User Interface} 
Our qualitative analysis of SHIELD's disruption predictions, focusing on real-world case studies (detailed in Appx.~\ref{app: case studies}), complements the quantitative findings and further illuminates the system's practical utility. A particularly salient example emerged in SHIELD's accurate prediction of a semiconductor shortage resulting from geopolitical tensions, made three weeks prior to widespread reporting. This early insight enabled proactive adjustments to procurement strategies, thereby demonstrating the system's considerable potential in mitigating complex supply chain risks.~We have developed an interactive user interface (Fig.~\ref{fig:user_interface}) for online disruption analysis.~This interface allows users to upload news report texts (Fig.~\ref{fig:user_interface part1 user input}), evaluate prediction scores, and edit visualization results for the final disruption analysis (Fig.~\ref{fig:user_interface part2 visualization and edit}). More details can be found in Appx.~\ref{app: Disruption Prediction User Interface Details}.

\subsection{Disruption Prediction Case Studies}
Our system effectively forecasted key supply chain disruptions, providing insights that enabled stakeholders to take proactive actions.
For example, following the passage of the Inflation Reduction Act (2022), which incentivized domestic EV battery production, our system predicted potential material shortages. By analyzing shifts in global material flows and the effects of policy changes on supply-demand dynamics, it enabled early interventions to minimize risks. Similarly, during geopolitical tensions between Australia and China in 2023, our system identified vulnerabilities in the lithium supply chain by monitoring export data and geopolitical developments, helping stakeholders adapt their sourcing strategies in time. In another instance, the system anticipated cobalt supply issues resulting from labor strikes and regulatory changes in the Democratic Republic of Congo (2023), allowing companies to diversify sources and increase inventory buffers.
These cases, detailed further in Appendix~\ref{app: case studies}, illustrate how data-driven predictions enhance supply chain resilience and support timely decision-making in a volatile global market.

\section{Conclusion}
\vspace{-2mm}
\label{sec:conclusion}
We present SHIELD, a two-stage framework that integrates Large Language Models~(LLMs) with domain expertise, yielding promising results in EV battery supply chain analytics and risk assessment. While demonstrating particular strength in early disruption detection and event prediction for critical battery materials, significant challenges remain in schema integration, real-time adaptability, and error reduction. Future research will systematically address these limitations, enhance the system's robustness, and explore broader applications across diverse industries and supply chain ecosystems.

\section*{Acknowledgments}
This research was supported by the Carnegie Mellon Manufacturing Futures Institute and the Manufacturing PA Innovation Program. The authors also thank the School of Computer Science (SCS) at Carnegie Mellon University, particularly the High Performance Computing (HPC), for providing essential computational resources.~The views expressed are those of the authors and do not necessarily reflect those of the funding agencies.

{
\bibliography{custom}
}

\clearpage
\newpage

\appendix

\section{Extended Related Work}
\noindent \textbf{Supply Chain Risk Management}.~AI techniques have been increasingly applied to predict and mitigate supply chain risks~\cite{ganesh2022future}.~While agent-based approaches~\cite{pino2010supply, giannakis2011multi, giannakis2016multi, blos2015application} enable inter-agent communication for forecasting, they often lack robust predictive capabilities and have limited parameter sets. Rule-based reasoning methods~\cite{gallab2019risk, behret2012fuzzy, paul2015supplier, paul2017quantitative, awasthi2018multi, camarillo2018knowledge} offer decision-making frameworks but provide minimal quantitative insights.~To address these limitations, Machine Learning~(DL) and Deep Learning~(DL) techniques have been employed~\cite{silva2017improving, hegde2020applications, garvey2015analytical, ruz2020sentiment, aljohani2023predictive, carbonneau2008application}, enhancing demand forecasting and disruption prediction~\cite{hendriksen2023artificial, makridis2023deep}. Recent studies have begun exploring the potential of large language models (LLMs) in supply chain management~\cite{ray2023leveraging}.~However, most current works prioritize predictive performance over interpretability, hindering practitioners' ability to make informed decisions.~Our approach addresses this gap by integrating LLMs for schema induction, extracting hierarchical knowledge-graph structures from academic resources to predict supply chain disruptions, thereby enhancing both predictive performance and interpretability.

\noindent \textbf{Schema Induction \& Learning}.~Building on foundational works~\cite{anderson1979general, evans1967brief}, recent advancements in language modeling have revolutionized schema induction. Large-scale language models~(LLMs)~\cite{brown2020language, rae2021scaling} have demonstrated remarkable capabilities in learning and generating schemas with minimal supervision.~Researchers have explored various strategies to enhance these models, including contextual explanations~\cite{wei2021finetuned, lampinen2022can}, rationale-augmented ensembles~\cite{wang2022rationale}, and incremental prompting~\cite{li2023open}. Transformer-based approaches~\cite{li2020connecting, li2021future} have proven particularly effective in managing schema generation for complex scenarios, representing schemas as graphs. Integrating human feedback~\cite{mondal2023interactiveie, yang2024human, zhang2023human} has been crucial in refining schema induction processes, addressing the limitations of automated methods.~Our approach leverages these advancements by employing an LLM-driven framework that integrates human feedback and expert knowledge into a human-in-loop system, thereby enhancing the practical accuracy and relevance of induced schemas.

\noindent \textbf{Event Extraction \& Analysis}.~Event extraction has evolved from manually crafted features~\cite{ahn2006stages} to neural models, including recurrent networks~\cite{nguyen2016joint, sha2018jointly}, convolutional networks~\cite{chen2015event}, graph networks~\cite{zhang2021abstract}, and transformers~\cite{liu2020event}. Recent research has focused on event argument extraction~\cite{wang2019hmeae} and explored zero-shot learning~\cite{huang2018zero} and weak supervision~\cite{chen2015event} to enhance performance.~Our approach incorporates various event extraction techniques, utilizing fine-tuned RoBERTa models and graph convolutional networks~(GCNs) to capture and analyze complex event relationships and their cascading impacts. This approach enables a deeper understanding of supply chain disruptions, distinguishing our system from traditional extraction techniques.

\section{Dataset}
\subsection{Schema Learning Dataset}
\label{app:schema_learning_dataset}
Our research began by examining the current state of EV batteries, focusing on the predominant types in use, such as lithium iron phosphate and nickel lithium batteries. We analyzed the battery production process and identified key raw materials, including lithium, cobalt, nickel, and graphite. Subsequently, we investigated the primary sources and production volumes of these materials. Through an extensive review of literature and statistical data, we categorized significant supply chain events into eight groups, three of which have long-term impacts. Each category was further divided into subcategories, and real-world events were identified to illustrate their impact on raw material supplies.

We also analyzed price trends for key raw materials over the past five years, using data from the London Metal Exchange (LME)\footnote{https://www.lme.com/en/}, to assess how news events influenced these prices. This research produced an initial scenario document listing the primary raw materials for EV batteries, their price trends, and an analysis of events causing supply chain issues and price fluctuations. Each category included at least one real-world example to demonstrate its impact.

The initial document was then submitted for review by a supply chain expert.~Based on the expert's feedback, we refined the events affecting the EV battery supply chain into 11 main categories, three with long-term impacts, and subdivided them into 27 subcategories.~Each subcategory was illustrated with 1-2 real-world events, and raw materials were further subdivided, such as different grades of nickel and types of lithium.~Categories with minimal impact were removed, resulting in a comprehensive and refined scenario document.

Based on the scenario document, we identified the raw materials and events related to the EV battery supply chain and began collecting an academic document dataset. Our data sources included Wikipedia entries, supply chain-related papers, and industrial reports.~After obtaining the raw data, we manually removed redundant information and noise, retaining only the paragraphs most relevant to the EV battery supply chain. Through meticulous organization, we compiled an academic document dataset consisting of 125 entries, distilled from 239 diverse sources: 200 academic papers, 22 industry reports, and 17 Wikipedia entries.~This curated dataset provides a focused knowledge base essential for analyzing and understanding the complexities of the EV battery supply chain.

\begin{figure*}[!htp]
\centering
    \includegraphics[width=1\textwidth]{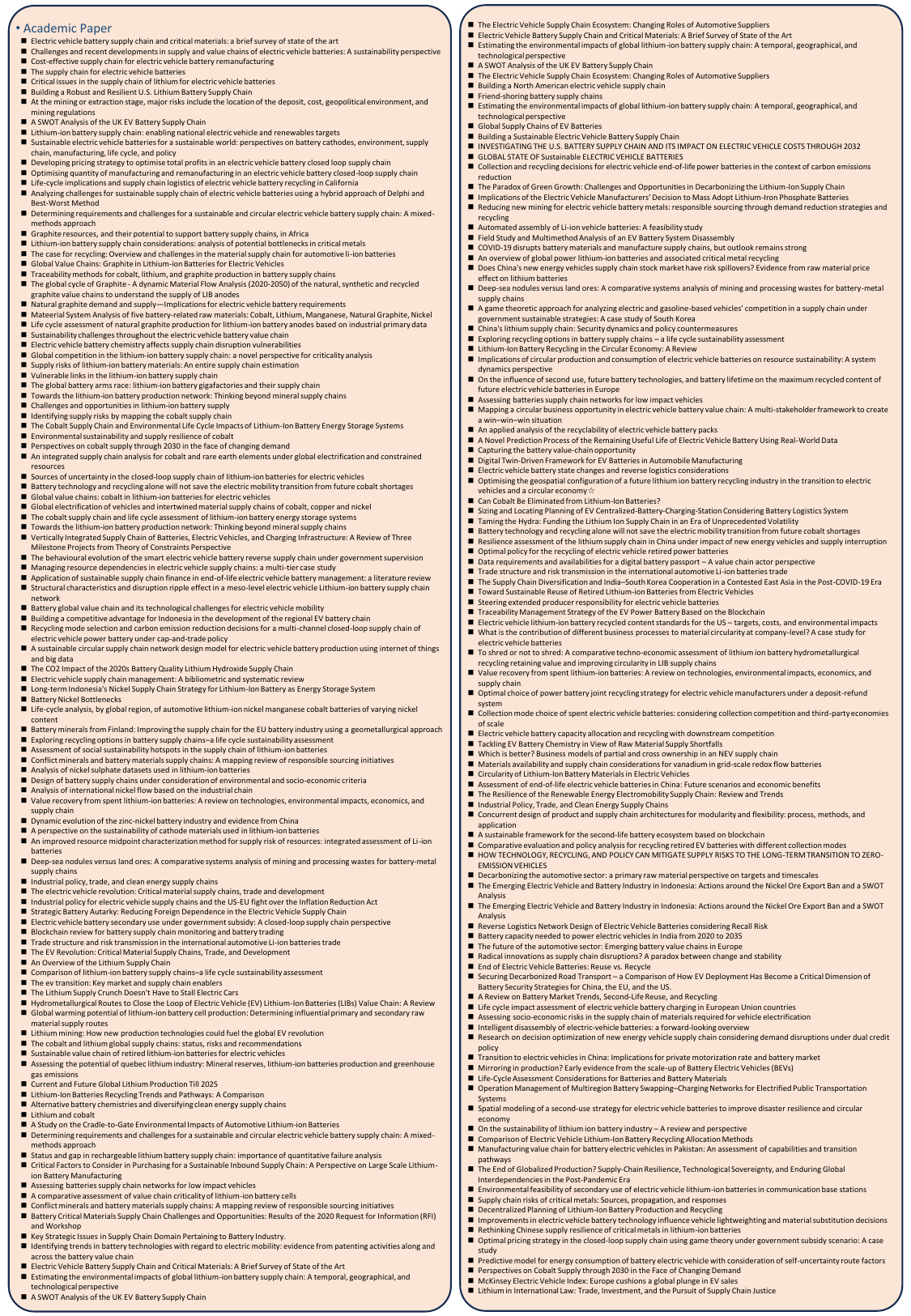}
\caption{Sources of academic papers.}
\label{app:academic_dataset_example}
\end{figure*}

\begin{figure*}[t]
\centering
    \includegraphics[width=1\textwidth]{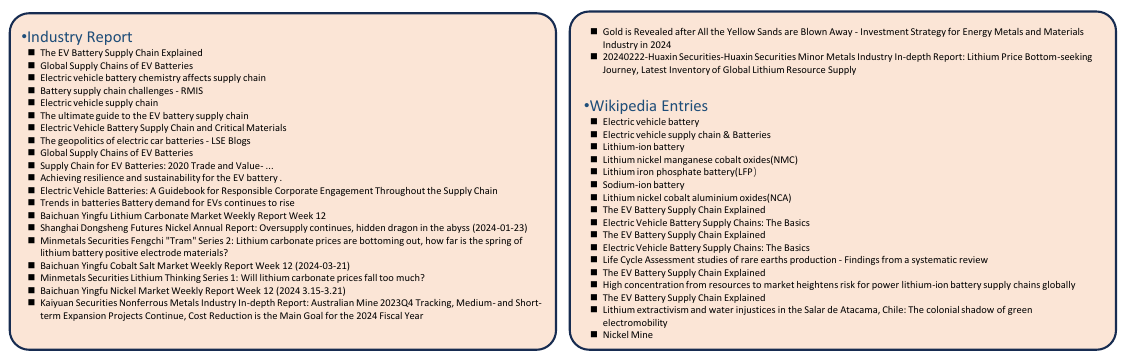}
\caption{Sources of Wikipedia entries and industry reports.}
\label{app:academic_dataset_example_wiki}
\end{figure*}

The resulting dataset encompasses a wealth of knowledge related to the EV battery supply chain, covering aspects such as raw material procurement, manufacturing processes, supply chain logistics, and market dynamics.
Table \ref{app:event_category} presents the event categories and example events. Events marked with * indicate potential long-term impacts, highlighting the various types of disruptions and their implications for the supply chain. Fig.~\ref{app:academic_dataset_example} and \ref{app:academic_dataset_example_wiki} illustrate the sources of the academic papers, Wikipedia entries, and industry reports used in compiling the dataset, demonstrating the breadth and diversity of our data sources.
By synthesizing this information, we aim to provide a robust foundation for understanding the complexities and challenges associated with the EV battery supply chain.

\subsection{Supply Chain News Dataset}
\label{app:newsdataset}
To comprehensively test our system, we constructed an EV Supply Chain News Dataset covering the period from January 2022 to December 2023. We initially developed a Python crawler using the \verb|requests| and \verb|BeautifulSoup| packages to scrape news titles and summaries from multiple websites, such as Google News\footnote{https://news.google.com/} and Infoplease\footnote{https://www.infoplease.com/}. This resulted in a collection of 643 records. To filter out news unrelated to the supply chain, we designed a prompt leveraging GPT-4o's language capabilities.~Using the summaries from the list, GPT-4o helped categorize events into various types, such as natural disasters, wars, trade policy, and political issues, tagging the relevant countries and regions.

Subsequently, we employed large language models (LLMs) to evaluate the relevance of each news event to the EV battery supply chain based on the following criteria, each worth 25 points:

\begin{enumerate}
    \item Whether natural disasters or humanitarian crises occurred in raw material production areas, such as China, Australia, Indonesia, Congo, Chile, Canada, or in EV production countries, such as China, Japan, South Korea, and the United States.
    \item Whether the event could affect trade relations in the aforementioned countries, including trade issues, sanctions, or wars.
    \item Whether the event could potentially disrupt international shipping routes due to conflicts or natural disasters near these routes.
    \item Whether the event is directly related to international trade.
\end{enumerate}

Events scoring below 25 points were initially eliminated, followed by a manual review of the remaining events, resulting in a refined list of 247 supply chain-related news events. The text data was sourced from reputable media outlets, including Reuters\footnote{https://www.reuters.com/}, BBC\footnote{https://www.bbc.com/}, and CNN\footnote{https://edition.cnn.com/}.~Additionally, to gather contemporaneous supply chain status information, we scraped company news and analysis reports from EV battery-related companies like Ford, Volkswagen, and CATL, as well as supply chain-related websites, totaling 118 reports. The raw data, including titles, publication dates, and content, was organized chronologically.

The raw data contained invalid information and advertisements, which were cleaned using regular expressions to remove most invalid information. We deployed Llama3-8b to filter out embedded advertisements, ensuring the dataset's purity and accuracy. After cleaning, irrelevant content was reduced by 15\%, and all data was systematically stored in a database, resulting in a refined meta dataset of 365 news documents.
The metadata contains approximately 152,000 words and 3,000 paragraphs. To validate our system's ability to detect connections between events, we randomly merged international news with contemporaneous corporate stories from the same quarter, creating 354 fused documents for a more diverse and challenging dataset. The final fusion dataset contains approximately 660,000 words and 12,000 paragraphs.

\begin{table*}[t]
\centering
\footnotesize
\caption{Event Categories and Example Events. Events marked with * indicate potential long-term impacts.}
\vspace{-2mm}
\label{app:event_category}
    \begin{tabular}{l l l}   
    \toprule
    Event Category & Subcategory & Example \\
    \midrule
    \multirow{2}{*}{Acquisition and Investment$^{\ast}$} & Investment from U.S. or Ally & U.S. invests in EV battery industry in Canada \\
     & Investment from Other Country & China invests in cobalt mines in DRC \\
     \midrule
    \multirow{2}{*}{Changes in Supply and Demand} & Demand Change & Demand for ore from the Philippines increases \\
     & Supply Change & Tight supplies of nickel ore in Indonesia \\
     \midrule
    \multirow{3}{*}{Enterprise Issue} & Production Halt or Reduction & Katanga halts cobalt mining \\
     & Enterprise Crisis & Katanga faces an equity crisis \\
     & Production Plan Adjustment & Kellyton Graphite increases production by 15\% \\
     \midrule
    \multirow{2}{*}{Economic Environment} & Macroeconomy & The U.S. and EU face continued inflation \\
     & Competition and Market Structure & Competition from China's low-priced EVs \\
     \midrule
    \multirow{2}{*}{EV Battery Technology Progress$^{\ast}$} & Product Upgrading & CATL releases Kirin battery \\
     & Production Technology Progress & Development of graphene batteries \\
     \midrule
    \multirow{3}{*}{Humanitarian and Ethical Crisis} & Forced Labor & Forced labor in production \\
     & Use of Child Labor & Child labor in cobalt mining in DRC \\
     & Human Rights Issue & Large numbers of refugees enter Europe \\
     \midrule
    \multirow{2}{*}{Natural Disaster} & Production Affected by Disaster & Australia floods affect lithium mining \\
     & Transportation Affected by Disaster & Tsunami destroys ports, disrupts shipping \\
     \midrule
    \multirow{4}{*}{Political Issue$^{\ast}$} & Regional Tension & Tensions between North and South Korea \\
     & Changes in International Relations & China's relations with the West deteriorate \\
     & Industry Nationalization & Nationalization of the lithium industry in Chile\\
     & Government Intervention & Europe promotes EVs for environmental reasons \\
    \midrule
    Sign a Supply Agreement & Sign a Supply Agreement & PE signs EV battery supply agreement with Tesla \\
    \midrule
    \multirow{3}{*}{Trade Policy} & Export Controls & China restricts graphite exports \\
     & Tax and Duties & China's tax rebates to EV companies \\
     & Trade Barriers & US tariffs on Chinese EV batteries \\
     \midrule
    \multirow{3}{*}{War and Conflict} & Internal Disorder or Rebellion & Civil unrest in Yemen \\
     & War Between Nations & Russo-Ukrainian War \\
     & Geopolitical Crisis & Houthi rebels attack merchant ship \\
    \bottomrule
    \end{tabular}
\end{table*}

\FloatBarrier

\begin{figure*}[!htp]
\centering
    \includegraphics[width=1\textwidth]{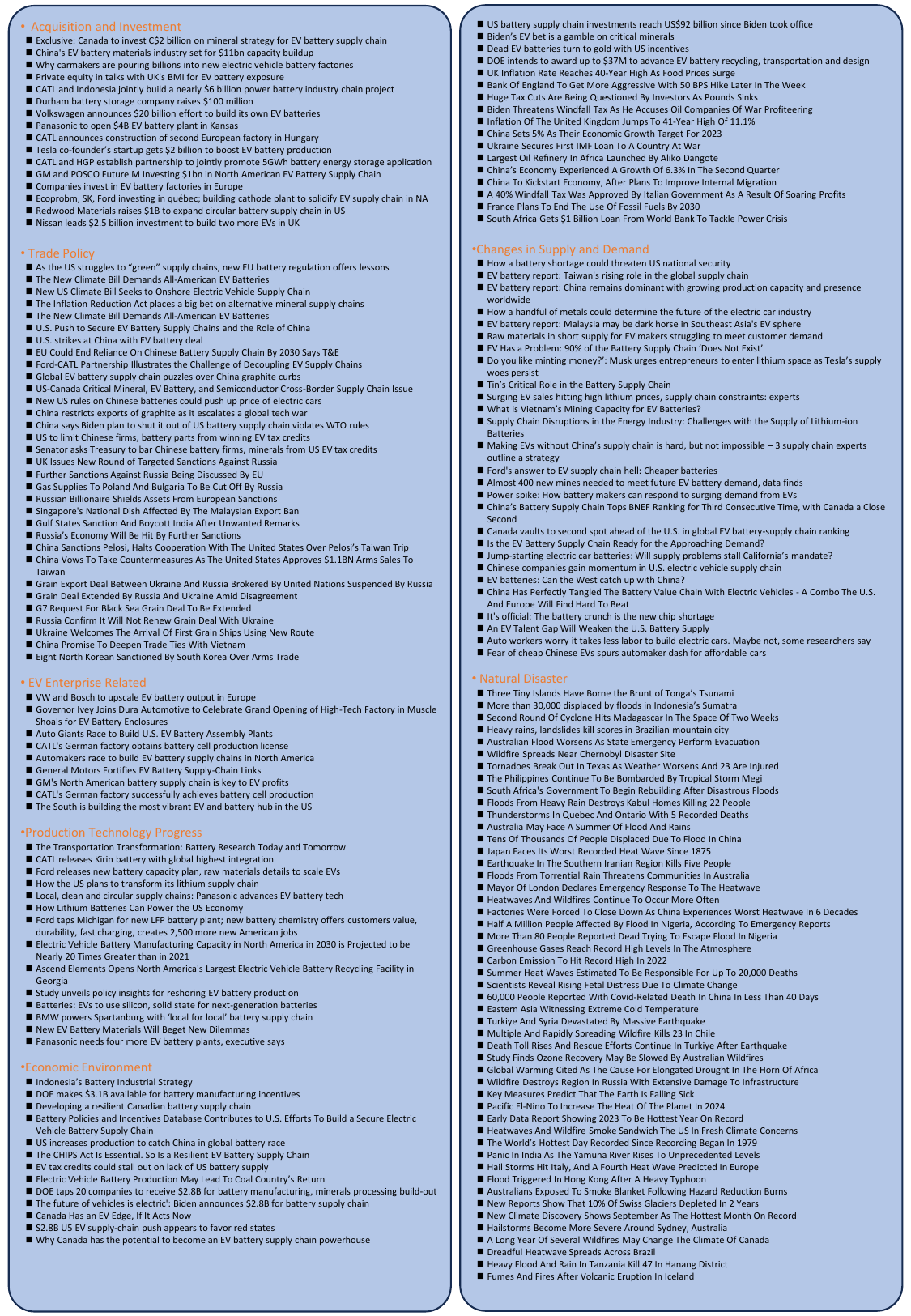}
\caption{Categories and examples of news articles in the dataset.}
\label{app:newsdataset_example}
\end{figure*}

\begin{figure*}[!htp]
\centering
    \includegraphics[width=1\textwidth]{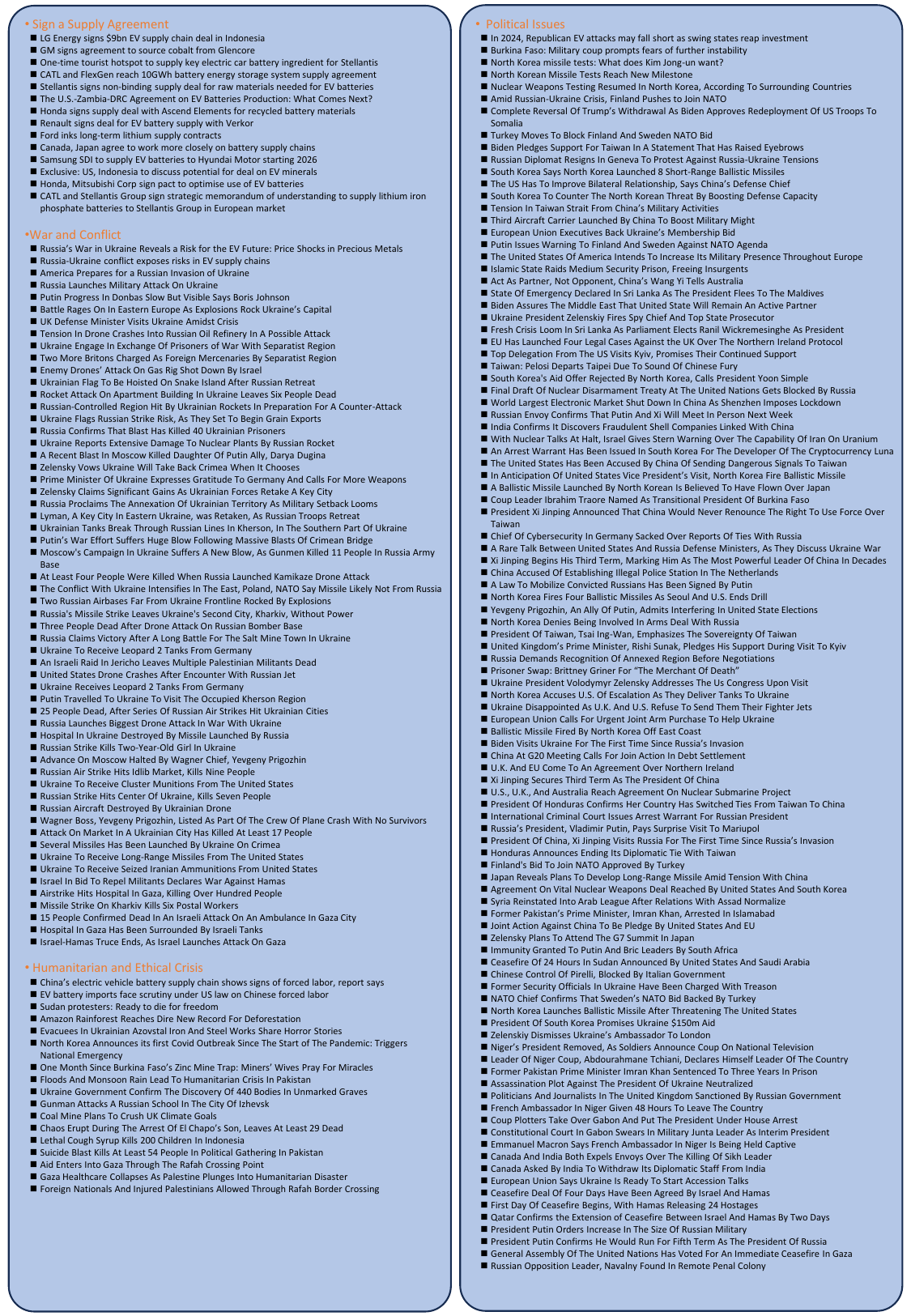}
\caption{Distribution of sources in the news dataset.}
\label{app:newsdataset_example_2}
\end{figure*}

Upon completing dataset collection, we conducted preliminary statistics and analysis on the news dataset.
Table \ref{app: event_category} shows the number of event types included in each quarter in the news dataset, providing a comprehensive overview of the various events tracked over time. Table \ref{app: news_wordcount} details the number of words and paragraphs in the dataset, highlighting the extensive scope of the collected data. Fig.~\ref{app:newsdataset_example} presents the categories and examples of news articles in the dataset, while Fig.~\ref{app:newsdataset_example_2} shows the distribution of sources in the news dataset, emphasizing the dataset's diversity and comprehensiveness.
The dataset covers global events that could impact the supply chain, such as natural disasters, trade issues, wars, enterprise issues, etc.

\begin{table*}[!ht]
    \centering
    \footnotesize
    \caption{The number of event types included in each quarter in the news dataset.}
    \vspace{-2mm}
    \begin{tabular}{l c c c c c c c c} 
        \toprule
        \textbf{Event Type}& \textbf{2022Q1} & \textbf{2022Q2} & \textbf{2022Q3} & \textbf{2022Q4} & \textbf{2023Q1}& \textbf{2023Q2} & \textbf{2023Q3} & \textbf{2023Q4} \\
        \midrule
        Acquisition and Investment & 0 & 5 & 4 & 0 & 2 & 2 & 2 & 1 \\
        Changes in Supply and Demand & 6 & 5 & 4 & 3 & 3 & 1 & 2 & 3 \\
        Enterprise Issue & 3 & 1 & 1 & 3 & 0 & 1 & 0 & 0 \\
        Economic Environment & 3 & 3 & 5 & 6 & 4 & 2 & 6 & 1 \\
        Humanitarian and Ethical Crisis & 1 & 5 & 3 & 1 & 2 & 0 & 2 & 3 \\
        Natural Disaster & 6 & 8 & 6 & 5 & 6 & 6 & 6 & 6 \\
        Political Issue & 3 & 14 & 18 & 16 & 14 & 10 & 15 & 10 \\
        EV Battery Technology Progress & 1 & 1 & 1 & 2 & 4 & 3 & 2 & 0 \\
        Sign a Supply Agreement & 0 & 3 & 1 & 1 & 2 & 2 & 1 & 4 \\
        Trade Policy & 2 & 6 & 5 & 2 & 5 & 1 & 2 & 9 \\
        War and Conflict & 3 & 6 & 12 & 10 & 5 & 7 & 7 & 7 \\
        \bottomrule
    \end{tabular}
\end{table*}

\begin{table*}[!ht]
    \centering
    \footnotesize
    \caption{The number of times each country is mentioned in the news dataset in each quarter.}
    \label{app: event_category}
    \vspace{-2mm}
    \begin{tabular}{l c c c c c c c c} 
        \toprule
        \textbf{Country}& \textbf{2022Q1} & \textbf{2022Q2} & \textbf{2022Q3} & \textbf{2022Q4} & \textbf{2023Q1}& \textbf{2023Q2} & \textbf{2023Q3} & \textbf{2023Q4} \\
        \midrule
        USA & 6 & 9 & 18 & 15 & 17 & 12 & 9 & 10 \\
        China & 1 & 10 & 12 & 6 & 12 & 4 & 4 & 9 \\
        EU & 3 & 11 & 6 & 4 & 8 & 4 & 8 & 2 \\
        Japan & 0 & 1 & 0 & 1 & 1 & 2 & 2 & 1 \\
        Russia & 4 & 10 & 4 & 14 & 6 & 6 & 4 & 3 \\
        Other & 20 & 42 & 38 & 33 & 31 & 23 & 35 & 40 \\
        \bottomrule
    \end{tabular}
\end{table*}

\begin{table}[!ht]
    \centering
    \footnotesize
    \caption{Statistics of the number of words and paragraphs in the dataset.}
    \label{app: news_wordcount}
    \vspace{-2mm}
    \begin{tabular}{l r r}   
        \toprule
         & Total Paragraphs & Total Words \\
         \midrule
        Meta Data & 3,022 & 152,489 \\
        Fusion Data & 12,070 & 660,054 \\
        \bottomrule
    \end{tabular}
\end{table}

\begin{figure*}[ht!]
\centering
\lstset{
    basicstyle=\ttfamily\footnotesize,
    breaklines=true,
    frame=single,
    captionpos=b,
    aboveskip=0pt,
    belowskip=0pt,
    columns=fullflexible,
    keepspaces=true,
    showspaces=false,
    showstringspaces=false,
    showtabs=false,
    tabsize=2
}
\begin{lstlisting}
According to the provided paragraphs:

### {}_Paragraphs_provided ###

extract a detailed hierarchical structure related to the EV battery supply chain. 
The hierarchical structure should include the following levels:
- **Event**: Anything that happens related to the EV battery supply chain.
- **Event ID**: A unique identifier for each event.
- **Description**: A detailed 2-3 sentence explanation of the event.
- **Participants**: All sub-events related to this event and their importance, the importance needs to be set as 0 ~ 1, the higher the more important.
- **Gate**: The relationship between an event and its sub-events:
  - Use **'and'** if no sub-events can be missing.
  - Use **'or'** if some sub-events can be missing.
  - Use **'xor'** if sub-events cannot exist simultaneously.
- **Relations**: The event-event relations (e.g., ev1.1>ev1.2, which means ev1.2 happens after ev1.1).
- If any level is empty, set its value to 'xxxx'.

Strictly use the exact following format for each event:
```
Event N
event: [Event Name]
event_id: evN
description: [Detailed Description]
participants: [Subevent 1] evN.1_P[Importance], [Subevent 2] evN.2_P[Importance], ...
Gate: [Gate]
Relations: [Event Relations]

Subevent N.1
subevent: [Subevent Name]
event_id: evN.1
description: [Detailed Description]
participants: [Subsubevent 1] evN.1.1_P[Importance], ...
Gate: [Gate]
Relations: [Event Relations]

Subsubevent N.1.1
subsubevent: [Subsubevent Name]
event_id: evN.1.1
description: [Detailed Description]
participants: [Subsubsubevent 1] evN.1.1.1_P[Importance], ...
Gate: [Gate]
Relations: [Event Relations]
```
\end{lstlisting}
\caption{Example of hierarchical structure extraction. (Part 1)}
\label{lst:example1}
\end{figure*}

\begin{figure*}[!htp]
\centering
\ContinuedFloat
\lstset{
    basicstyle=\ttfamily\footnotesize,
    breaklines=true,
    frame=single,
    captionpos=b,
    aboveskip=0pt,
    belowskip=0pt,
    columns=fullflexible,
    keepspaces=true,
    showspaces=false,
    showstringspaces=false,
    showtabs=false,
    tabsize=2
}
\begin{lstlisting}
Use the provided example for guidance: 
### Example:
**Input Paragraph**:
```
Three main methods are used in lithium-ion recycling: pyrometallurgical, hydrometallurgical, bioleaching, and direct recycling. The battery is melted in a hot furnace to recover some of the cathode metal in pyrometallurgy. Pyrometallurgy employs extreme heat to transform metal oxides into cobalt, copper, iron, and nickel alloys. Although it has a straightforward process and a reasonably mature technology, the main drawbacks are its high cost and high environmental pollution. Hydrometallurgy is a metal recovery method involving aqueous solutions to perform leaching processes to precipitate a particular metal. In hydrometallurgy, specialized solution reagents are primarily used to leach the targeted metals out from the cathode substance. Although it is a highly effective and power-efficient method, its drawbacks include a lengthy production time and a complicated process. Combinations of both pyrometallurgy and hydrometallurgy are also used due to their advantages in sorting starting materials for cells. The bioleaching technique uses bacteria to retrieve precious metals, but it is challenging because the bacteria need a substantial amount of time to grow and are easily susceptible to contamination.
```

**Extracted Hierarchical Structure**:

```
Event 1
event: lithium-ion recycling
event_id: ev1
description: Methods for recycling lithium-ion batteries including pyrometallurgical, hydrometallurgical, bioleaching, and direct recycling.
participants: pyrometallurgical ev1.1_P1, hydrometallurgical ev1.2_P1, bioleaching ev1.3_P1
Gate: or
Relations: ev1.1>ev1.3, ev1.2>ev1.3

Subevent 1.1
subevent: pyrometallurgical
event_id: ev1.1
description: Employs extreme heat to transform metal oxides into cobalt, copper, iron, and nickel alloys.
participants: metal oxides ev1.1.1_P1, cobalt ev1.1.2_P0.5, copper ev1.1.3_P0.5, iron ev1.1.4_P0.5, nickel alloys ev1.1.5_P0.5
Gate: and
Relations: ev1.1.1>ev1.1.2, ev1.1.1>ev1.1.3, ev1.1.1>ev1.1.4, ev1.1.1>ev1.1.5

Subevent 1.2
subevent: hydrometallurgy
event_id: ev1.2
description: Uses aqueous solutions to leach targeted metals out from the cathode substance.
participants: xxxx
Gate: xxxx
Relations: xxxx

Subevent 1.3
subevent: bioleaching
event_id: ev1.3
description: Uses bacteria to retrieve precious metals.
participants: xxxx
Gate: xxxx
Relations: xxxx
```

Think about this extracted structure step by step:
Starting with the first sentence in the paragraph 'Three main methods are used in lithium-ion recycling: pyrometallurgical, hydrometallurgical, bioleaching, and direct recycling.' From this sentence, we learn that 'pyrometallurgical', 'hydrometallurgical', 'bioleaching, and direct recycling' are three methods of 'lithium-ion recycling', so select 'lithium-ion recycling' as the event, and the three methods as subevents and participants of 'lithium-ion recycling'.
\end{lstlisting}
\caption{Example of hierarchical structure extraction. (Part 2)}
\label{lst:example2}
\end{figure*}

\section{Hierarchical Structure Extraction}
\label{app: prompt hierarchical}
We utilize large language models (LLMs) to extract hierarchical structures ($\mathbf{H}$) that capture main events ($\mathbf{E}$) and sub-events ($\mathbf{E_{sub}}$) based on our prompt, as illustrated in Fig.~\ref{lst:example1}. 

In a hierarchical structure ($\mathbf{H}$):
\begin{itemize}
    \item An \textit{event} ($\mathbf{E}$) refers to anything that happens related to the EV battery supply chain. There can be multiple events $\left\langle \mathbf{E}_{1}, \mathbf{E}_{2}, \ldots, \mathbf{E}_{n} \right\rangle$ in one hierarchical structure $\mathbf{H}$.
     \item An \textit{event\_id} is a unique identifier code assigned for each specific event.
    \item A \textit{description} provides a detailed 2-3 sentence textual explanation of the event.
    \item \textit{Participants} include all sub-events ($\mathbf{E_{sub}}$) related to the main event, and a \textit{subsubevent} ($\mathbf{E}_{subsub}$) can be used if an event is part of a sub-event within the hierarchy.
\end{itemize}

The suffix \textit{P0.5} indicates the importance of a sub-event to its parent event. The \textit{Gate} specifies the relationship between the main event and its sub-events:
\begin{itemize}
    \item Use 'and' if no sub-events can be missing.
    \item Use 'or' if some sub-events can be missing.
    \item Use 'xor' if sub-events cannot exist simultaneously.
\end{itemize}

\textit{Relations} describe the connections between events. For example, if \textit{ev1.2} is caused by (happens after) \textit{ev1.1}, it is expressed as 'ev1.1>ev1.2'.

Our prompt includes demonstration and chain of thought (CoT) techniques:
\begin{itemize}
    \item We manually annotated the hierarchical structure for one text in the schema learning dataset to use as an example in the prompt.
    \item We provided a step-by-step CoT, showing how $\mathbf{E}$ and $\mathbf{E_{sub}}$ in $\mathbf{H}$ were extracted from specific sentences in the schema learning dataset.
\end{itemize}

The prompt given to the LLMs is detailed and specific, ensuring that the models understand the exact format and type of information we are extracting. By integrating demonstration and CoT techniques, our prompt provides clear guidance to the LLMs, improving the accuracy and relevance of the extracted structures.~Below is an example of the prompt used in Fig.~\ref{lst:example1}.

To validate our approach, we tested the prompt with various texts from the schema learning dataset. The hierarchical structures extracted were compared with manually annotated structures to ensure accuracy and consistency. This process ensured that the LLMs reliably produced high-quality hierarchical structures that aligned with expert knowledge in the EV battery supply chain domain.

\section{Schema Generation \& Merging}
\label{app: schema merging}
With human-in-the-loop schema induction, our schema learning dataset generated 125 individual schemas $\left\langle \mathbf{S}_{1}, \mathbf{S}_{2}, \ldots, \mathbf{S}_{125} \right\rangle$.~To create a single comprehensive schema, it is essential to merge all individual schemas into a final schema ($\mathbf{S_{\text{final}}}$). The process of merging schema format files involves systematically integrating multiple schemas into a cohesive schema. The key components include \textit{context}, \textit{id}, \textit{events}, and \textit{relations}. These components determine the information in each event and its correlation with other events, hence the merging process must address all of them.

\begin{algorithm}[!ht]
\caption{Schemas Merging Pseudocode}
\label{alg:schema_merging}
\footnotesize
\begin{algorithmic}[1]
    \State \textbf{Input:} List of schemas
    \State \textbf{Output:} Merged schema

    \State \textbf{1. Merge contexts from all schemas:}
    \ForAll{schema \textbf{in} schemas}
        \ForAll{context \textbf{in} schema["@context"]}
            \If{context \textbf{not in} merged\_contexts}
                \State Add context to merged\_contexts
            \EndIf
        \EndFor
    \EndFor

    \State \textbf{2. Merge events from all schemas by event name:}
    \ForAll{schema \textbf{in} schemas}
        \ForAll{event \textbf{in} schema["events"]}
            \State event\_name = event["name"]
            \If{event\_name \textbf{not in} merged\_events}
                \State merged\_events[event\_name] = event
            \Else
                \State merged\_events[event\_name] = merge\_event\_details(merged\_events[event\_name], event)
            \EndIf
        \EndFor
    \EndFor

    \State \textbf{3. Merge relations / update event IDs by event names:}
    \ForAll{schema \textbf{in} schemas}
        \ForAll{relation \textbf{in} schema["relations"]}
            \State subject\_name = GET event name by event ID relation["relationSubject"]
            \State object\_name = GET event name by event ID relation["relationObject"]
            \If{subject\_name \textbf{in} name\_to\_id \textbf{and} object\_name \textbf{in} name\_to\_id}
                \State relation["relationSubject"] = name\_to\_id[subject\_name]
                \State relation["relationObject"] = name\_to\_id[object\_name]
                \If{relation \textbf{not in} merged\_relations}
                    \State Add relation to merged\_relations
                \EndIf
            \EndIf
        \EndFor
    \EndFor

    \State \textbf{4. Final Schema:}
    \State The final merged schema includes all merged \textit{contexts}, \textit{events}, and \textit{relations}, and is saved for evaluation.
\end{algorithmic}
\normalsize
\end{algorithm}

To begin the merging process, we first aggregate the \textit{context} data from all schemas. Each \textit{context} is added to a \textit{merged\_contexts\_list}, ensuring that duplicate contexts are avoided. This step is crucial to maintain a unified context for the merged schema.
Next, we proceed to merge events from all schemas. Using the event name as the identifier, we check if the event already exists in the \textit{merged\_events\_list}. If the event exists, its details are merged with the existing event; otherwise, the event is added directly to the list. This ensures that all events are comprehensively integrated without duplication.

Following the merging of events, we then merge relations and update event IDs. This involves retrieving the event names from the event IDs for \textit{relationSubject} and \textit{relationObject} and updating the relations accordingly. It is important to ensure that both \textit{subject\_name} and \textit{object\_name} are present in the \textit{name\_to\_id} dictionary, which stores event names and their related event IDs. The updated relation is added to the \textit{merged\_relations} list if it is not already present, ensuring all connections are accurately maintained.

Finally, the comprehensive merged schema ($\mathbf{S_{final}}$) is created by including all merged \textit{contexts}, \textit{events}, and \textit{relations}. The detailed pseudocode for merging schemas is shown in Algorithm~\ref{alg:schema_merging}. This algorithm ensures that all relevant information is retained and accurately integrated, resulting in a comprehensive schema that encapsulates the full breadth of the data from the schema learning dataset. The final schema ($\mathbf{S_{final}}$) enables accurate and efficient knowledge extraction and organization, enhancing the utility of the dataset for downstream tasks such as event prediction and analysis.

\section{Schema Management System}
\label{app: hmi interface}
The schema management interface (Fig.~\ref{app:fig:schema_management_system}) facilitates the visualization, editing, and management of schemas. It includes the following modules:

\begin{figure*}[!ht]
  \centering
    \includegraphics[width=\linewidth]{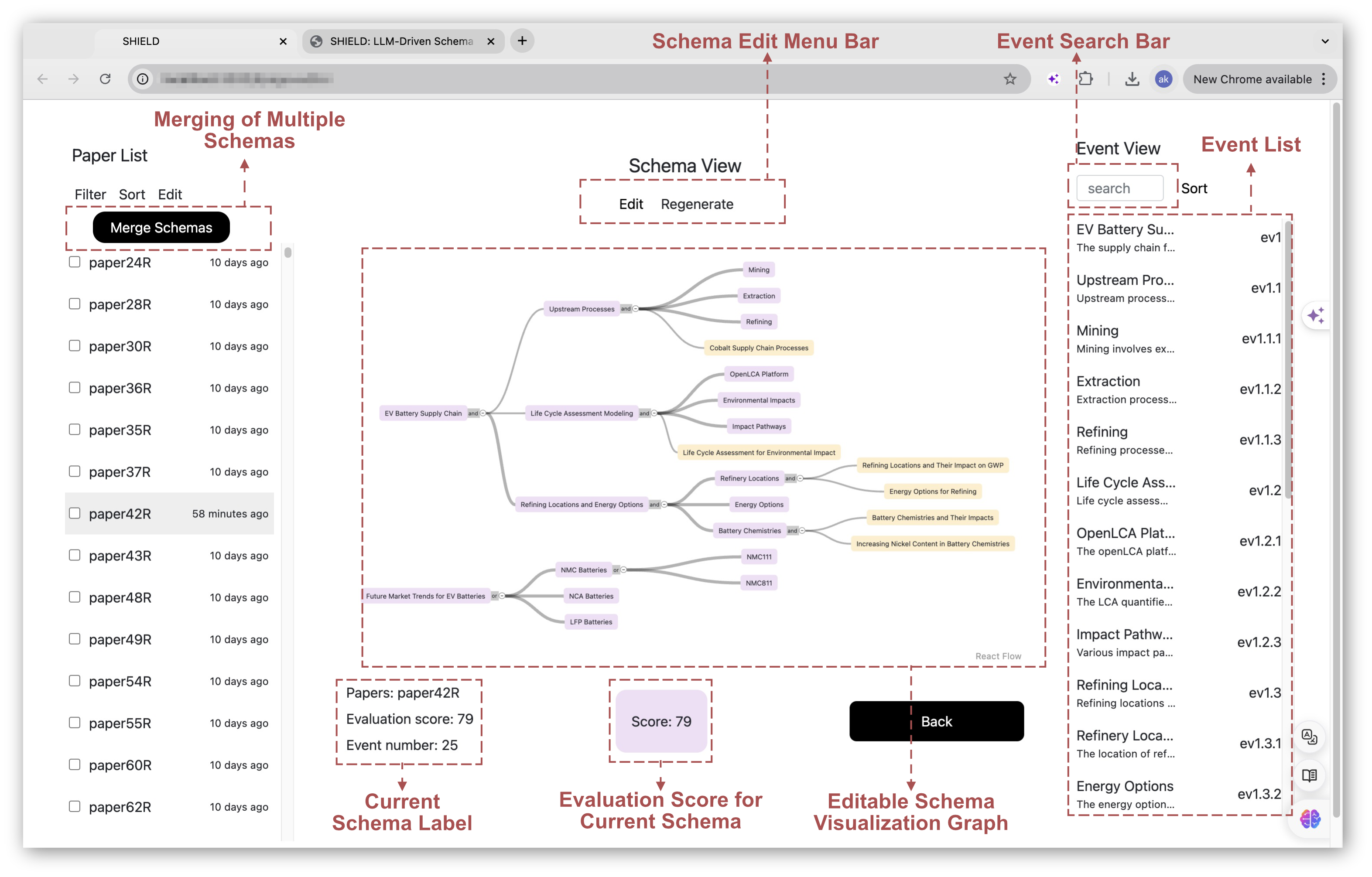}
    \vspace{-2mm}
    \caption{User interface for our schema management system.}
    \vspace{-4mm}
    \label{app:fig:schema_management_system}
\end{figure*}

\subsection{Schema Viewer}
The schema viewer is crucial for visualizing schemas, providing an intuitive representation of events.~It organizes events into a left-to-right tree structure, highlighting parent-child relationships. Within this structure, before-after relationships among child nodes are indicated through arrows and vertical ordering.~Users can expand event nodes to reveal details such as descriptions, importance levels, and participant roles.

Key features of the schema viewer include:
\begin{itemize}
    \item \vspace{-0.5em} \textbf{Interactive Exploration:} Users can click on nodes to expand or collapse details about events and sub-events.
    \item \vspace{-0.5em} \textbf{Contextual Information:} Hovering over a node displays additional context and metadata associated with the event.
    \item \vspace{-0.5em} \textbf{Dynamic Layout:} The tree structure dynamically adjusts to accommodate the addition or removal of nodes, maintaining a clear and organized visual representation.
   \item \vspace{-0.5em} \textbf{Collapsible Subtrees:} Users can collapse and expand subtrees to manage large schemas.
    \item \vspace{-0.5em} \textbf{Search Functionality:} A search bar allows users to quickly locate specific events or entities within the schema.
    \item \vspace{-0.5em} \textbf{Real-Time Data Binding:} The viewer updates in real-time as changes are made, ensuring the displayed schema is always current.
    \item \vspace{-0.5em} \textbf{Highlighting and Filtering:} Users can highlight specific paths or filter events based on criteria such as importance or type.
\end{itemize}

\subsection{Schema Editor}
The schema editor allows users to interactively modify schemas. Users can add, edit, and delete events, sub-events, and relationships within the schema. Key functionalities include:
\begin{itemize}
    \item \vspace{-0.5em} \textbf{Drag-and-Drop Interface:} Users can drag and drop nodes to reassign parent-child relationships or reorder events.
    \item \vspace{-0.5em} \textbf{Form-Based Editing:}~Clicking on a node opens a form where users can edit event details, such as descriptions, importance levels, and participant roles.
    \item \vspace{-0.5em} \textbf{Validation Checks:} The editor performs real-time validation to ensure that all changes adhere to the schema format and constraints.
    \item \vspace{-0.5em} \textbf{Undo/Redo Features:}~Users can easily undo or redo changes to maintain the integrity of the schema editing process.
    \item \vspace{-0.5em} \textbf{Schema Versioning:}~The editor maintains different versions of schemas, allowing users to track changes over time and revert to previous versions if necessary.
    \item \vspace{-0.5em} \textbf{Bulk Operations:} Users can perform bulk operations such as adding multiple events or updating several nodes at once.
    \item \vspace{-0.5em} \textbf{Conflict Resolution:}~The editor provides tools to resolve conflicts when multiple users make changes simultaneously.
\end{itemize}

\subsection{Frontend Architecture}
The frontend of system is implemented as a single-page web application using React and TypeScript. This setup connects to an API server that provides application logic and access to a centralized schema database. The use of a browser-based application offers several advantages, including no need for user installations, centralized data management, and extensive functionality through JavaScript libraries.~Key components include:
\begin{itemize}
    \item \vspace{-0.5em} \textbf{React\footnote{https://react.dev/}:} A JavaScript library for building user interfaces, providing the foundation for the application's dynamic and responsive design.
    \item \vspace{-0.5em} \textbf{TypeScript\footnote{https://www.typescriptlang.org/}:} A statically typed superset of JavaScript, enhancing code reliability and maintainability.
    \item \vspace{-0.5em} \textbf{GoJS\footnote{https://gojs.net/latest/index.html}:} A JavaScript library for creating interactive diagrams, enabling robust schema visualization.
    \item \vspace{-0.5em} \textbf{API Integration:} The frontend communicates with the backend through API calls, fetching and submitting schema data.
    \item \vspace{-0.5em} \textbf{Responsive Design:} The application is optimized for various screen sizes and devices, ensuring usability across different platforms.
    \item \vspace{-0.5em} \textbf{State Management:} The application uses state management libraries such as Redux to manage and synchronize the state of the schema data across different components.
    \item \vspace{-0.5em} \textbf{Performance Optimization:}~Techniques such as code splitting and lazy loading are employed to ensure fast load times and smooth interactions.
\end{itemize}

The client-side application requests Schema Definition Files (SDF) from the API server and displays them to users. Edits to the SDF are maintained locally until the user saves the changes, synchronizing the server-side copy with the client's modifications. A simple locking mechanism is employed to prevent simultaneous edits by multiple users on the same schema, ensuring data integrity.

\subsection{Backend Architecture}
The backend of the interface is developed in Python, leveraging the Falcon web server framework, served by Gunicorn and nginx, and supported by a SQLite database. The backend is designed to be lightweight, minimalist, and easy to comprehend. Most functionalities are concentrated in the frontend to maintain responsiveness and interactivity, allowing the backend to focus primarily on data management. Python's versatility and popularity make it a suitable choice for the dynamic requirements of the system. Static typing in Python is enforced using Mypy\footnote{https://mypy-lang.org/} to facilitate development and reduce trivial bugs.~Key components include:
\begin{itemize}
    \item \vspace{-0.5em} \textbf{Falcon\footnote{https://falcon.readthedocs.io/}:} A minimalist web framework for building high-performance APIs, facilitating efficient communication between the frontend and backend.
    \item \vspace{-0.5em} \textbf{Gunicorn\footnote{https://gunicorn.org/}:} A Python WSGI HTTP server for running web applications, ensuring robust and scalable performance.
    \item \vspace{-0.5em} \textbf{nginx\footnote{https://nginx.org/en/}:} A high-performance web server and reverse proxy, providing load balancing and enhancing security.
    \item \vspace{-0.5em} \textbf{SQLite\footnote{https://www.sqlite.org/}:}~A lightweight, disk-based database, chosen for its simplicity and reliability.
    \item \vspace{-0.5em} \textbf{RESTful API\footnote{https://restfulapi.net/}:} The backend exposes a RESTful API for the frontend to interact with schema data, supporting CRUD operations.
    \item \vspace{-0.5em} \textbf{Security Features:} Implementations such as HTTPS, authentication, and authorization to ensure data privacy and integrity.
    \item \vspace{-0.5em} \textbf{Scalability:} The architecture is designed to scale horizontally, with load balancers and database replication as needed.
\end{itemize}

\subsection{AI-Driven Suggestions}
The interface incorporates AI-driven suggestions to assist users in schema creation and modification. Large Language Models (LLMs) analyze existing schemas and user inputs to provide recommendations for schema elements, relationships, and structures. These suggestions are presented in real-time, enhancing user productivity and ensuring the creation of accurate and comprehensive schemas.

Key features of AI-driven suggestions include:
\begin{itemize}
    \item \vspace{-0.5em} \textbf{Contextual Recommendations:} The system provides context-aware suggestions based on the current schema and user actions.
    \item \vspace{-0.5em} \textbf{Smart Auto-Completion:} As users type or modify schema elements, the interface offers auto-completion options to expedite the editing process.
    \item \vspace{-0.5em} \textbf{Error Detection:} The AI models detect potential errors or inconsistencies in the schema and suggest corrections.
    \item \vspace{-0.5em} \textbf{Learning from User Feedback:} The AI models improve over time by learning from user feedback and interactions, refining their suggestions and increasing accuracy.
    \item \vspace{-0.5em} \textbf{Interactive Tutorials:} The interface includes tutorials and guidance to help users understand and leverage AI-driven suggestions effectively.
\end{itemize}

\section{Details of Event Extraction}
\label{app: event extraction}

\subsection{Event Span Identification}
Event span identification involves locating and marking the spans of events within input text. We use two models for this task:

\textbf{Base Model:} This model is a fine-tuned version of the RoBERTa-large language model \cite{liu2019roberta}, trained on an internally annotated dataset. The task is formulated as sequence tagging, where the model identifies the start and end positions of event spans. For instance, in the context of supply chain disruptions, the model identifies spans corresponding to events like factory shutdowns, transport delays, or material shortages. This aligns with the cross-sentence event detection described in the main text:
\begin{equation}
\text{EventDetect}_{\text{multi-sentence}}(\mathbf{T}) \rightarrow \mathbf{E_C}
\end{equation}
where \(\mathbf{T}\) represents the input text and \(\mathbf{E_C}\) the detected events. The model uses contextual information from neighboring sentences to accurately detect event boundaries, ensuring that even complex events spanning multiple sentences are correctly identified.

\textbf{Guided Model:} Inspired by \citet{wang2021query}, this model uses a query-based approach to focus on schema-related events. The process involves two stages as follows:
\begin{enumerate}
    \item \textbf{Discriminator Stage:} Queries representing event types are paired with sentences to predict if the query corresponds to an event type mentioned in the sentence. For example, queries include "factory shutdown due to labor strike" or "delay in shipping materials." This stage helps in filtering sentences that are likely to contain relevant events.
    \item \textbf{Span Extraction Stage:} Sentences identified in the discriminator stage are further processed to extract event spans using sequence tagging. This ensures that the extracted spans are relevant to the supply chain context. By using sequence tagging, the model accurately marks the start and end points of events within the identified sentences.
\end{enumerate}
This approach supports the cross-sentence event detection described in the main text, enriching event spans with relevant context and ensuring high precision in event identification.

\subsection{Event Argument Extraction}
Event argument extraction involves identifying the roles and participants associated with events. This task is framed as extractive question answering, where the model extracts argument spans from the text based on role-specific questions. We fine-tune RoBERTa-large~\cite{liu2019roberta} on our internally annotated dataset with a sequence tagging loss function. For supply chain disruptions, arguments might include the specific factories, transportation modes, or materials directly affected by the event. 

The extraction process is as follows:
\begin{enumerate}
    \item \textbf{Role-Specific Questions:} The model is trained to answer questions like "Which factory was shut down?" or "What material was delayed?" This method ensures that the arguments are specific and relevant.
    \item \textbf{Contextual Embeddings:} This step is enriched by contextual embeddings generated by BERT:
    \begin{equation}
    \text{BERT}_{\text{context}}(\mathbf{E_C}) \rightarrow \mathbf{C_E}
    \end{equation}
    generating contextual embeddings \(\mathbf{C_E}\). These embeddings provide rich semantic information, enabling the model to better understand the context and improve the accuracy and relevance of the extracted arguments.
\end{enumerate}

\subsection{Time Expression Linking \& Normalization}
Time expression linking connects time expressions to their corresponding events. Similar to argument extraction, this task uses extractive question answering to find start and end times for events. We fine-tune RoBERTa-base using the TempEval3 dataset \cite{uzzaman2012tempeval}. 

The process includes:
\begin{enumerate}
    \item \textbf{Extraction:} The model identifies time expressions within the text and links them to the corresponding events, ensuring that the timeline of events is accurately captured.
    \item \textbf{Normalization:} Identified time expressions are then normalized into standard formats using SUTime \cite{chang2012sutime} and HeidelTime \cite{strotgen2013multilingual}. For example, expressions like "next Monday" are converted into specific dates.
\end{enumerate}
For supply chain disruptions, this ensures that timelines for events like "shipment delayed from March 15 to March 20" are accurately captured. This integrates into the event parameter extraction process, ensuring coherence and consistency.

\subsection{Event Temporal Ordering}
Event temporal ordering determines the chronological sequence of events. We frame this task as extractive question answering to address label imbalance issues, fine-tuning RoBERTa-large~\cite{liu2019roberta} with a sequence tagging loss. 

Steps include:
\begin{enumerate}
    \item \textbf{Pairwise Temporal Relations:} The model identifies pairwise temporal relations between events, such as "Event A happened before Event B."
    \item \textbf{Consistency Checking:} Pairwise temporal relations are processed using Integer Linear Programming (ILP)~\cite{schrijver1998theory} to ensure consistency and resolve any conflicts. This method helps in constructing a coherent timeline of events.
\end{enumerate}
This is crucial for understanding the sequence of disruptions in supply chains, such as how a factory shutdown leads to delayed shipments. This aligns with the logical constraints and argument coreference to maintain event relationships modeled by GCNs:
\begin{equation}
\text{LogicCoref}(\mathbf{P_C}) \rightarrow \mathbf{P_F}
\end{equation}

\subsection{Coreference Resolution}
We perform both within-document and cross-document coreference resolution using models fine-tuned on datasets like OntoNotes 5.0~\cite{pradhan2013towards}. 

The resolution process involves:
\begin{enumerate}
    \item \textbf{Entity Clustering:} Entity and event coreference clusters are identified and linked to ensure consistency across documents. This helps in tracking the same entities and events mentioned in different parts of the text.
    \item \textbf{Cross-Document Linking:} Linking entities and events across multiple documents ensures that all references to a specific factory, supplier, or shipment are recognized as the same entity.
\end{enumerate}
This is critical for tracking entities like factories, suppliers, and shipments across multiple reports of supply chain disruptions. This supports the coreference resolution and event linking described in the main text:
\begin{equation}
\text{CorefLink}(\mathbf{E_C}) \rightarrow \mathbf{E_L}
\end{equation}
yielding linked events \(\mathbf{E_L}\).

\subsection{Graph Convolutional Networks (GCNs) for Event Relationship Modeling}
We leverage Graph Convolutional Networks (GCNs) to model complex event relationships and assess each event's impact. This involves constructing a graph where events are nodes and their interactions are edges.

Steps include:
\begin{enumerate}
    \item \textbf{Node Importance Calculation:} Each node's importance is calculated using centrality measures, such as degree centrality, betweenness centrality, and eigenvector centrality. These measures help in understanding the influence of each event within the network.
    \item \textbf{Edge Impact Calculation:} Edges represent the magnitude of impact, quantified by measures such as event severity and frequency of occurrence.
\end{enumerate}
The impact score is then calculated as:
\begin{equation}
\text{ImpactScore}(e_i) = \text{Centrality}(e_i) + \text{Magnitude}(e_i)
\end{equation}
where \(\text{Centrality}(e_i)\) reflects the event's importance within the network, and \(\text{Magnitude}(e_i)\) quantifies the event's impact intensity.

\subsection{Logical Constraints and Argument Coreference}
To ensure the robustness of our event extraction pipeline, we apply logical constraints and argument coreference resolution.

This involves multiple steps to refine the extracted event parameters and ensure logical consistency:

\textbf{Logical Constraints Application:}
\begin{enumerate}
    \item \textbf{Defining Logical Rules:} We define a set of logical rules to maintain consistency within the extracted events. These rules include:
    \begin{itemize}
        \item \textit{Temporal constraints}: An event must occur before another if there is a chronological dependency.
        \item \textit{Causal relationships}: If Event A causes Event B, then Event A must be identified as a precursor to Event B.
    \end{itemize}
    \item \textbf{Implementation:} The defined logical rules are implemented using a logic-based reasoning system that checks for any violations and rectifies them. For instance, if an event is detected as occurring before its cause, the system flags this inconsistency and corrects the sequence.
\end{enumerate}

\textbf{Argument Coreference Resolution:}
\begin{enumerate}
    \item \textbf{Coreference Detection:} We identify coreferences within and across documents. This involves detecting instances where different expressions refer to the same entity or event.
    \begin{itemize}
        \item \textit{Within-Document Coreference:} Ensures that all mentions of an entity within a single document are linked.
        \item \textit{Cross-Document Coreference:} Links mentions of the same entity or event across multiple documents to ensure global consistency.
    \end{itemize}
    \item \textbf{Refinement Process:}
    \begin{itemize}
        \item \textit{Cluster Formation:} Entities and events identified as coreferent are grouped into clusters.
        \item \textit{Coreference Chains:} We create chains of coreferent mentions, which are used to refine event parameters and ensure that all related mentions are consistently linked.
        \item \textit{Manual Verification:} After automatic coreference resolution, manual verification is performed by domain experts to ensure accuracy and address any ambiguities.
    \end{itemize}
\end{enumerate}

\textbf{Combining Logical Constraints \& Coreference:}
\begin{enumerate}
    \item \textbf{Integration:} The logical constraints and coreference resolution processes are integrated to produce a coherent and logically consistent set of event parameters:
    \begin{equation}
    \text{LogicCoref}(\mathbf{P_C}) \rightarrow \mathbf{P_F}
    \end{equation}
    \item \textbf{Validation:} The final set of event parameters \(\mathbf{P_F}\) undergoes a validation process to ensure that all logical rules and coreference chains are satisfied. This step is crucial for maintaining the integrity of the event extraction pipeline.
    \item \textbf{Feedback Loop:} A continuous feedback loop is established where the output is reviewed and refined based on new data and expert feedback. This iterative process helps in improving the model's performance over time.
\end{enumerate}

By applying these detailed logical constraints and advanced coreference resolution techniques, we ensure that the event extraction pipeline produces high-quality, reliable, and contextually accurate event data, which is essential for robust supply chain disruption analysis.

\section{Details of Event Matching \& Instantiation}
\label{app: event matching instantiation}

Event matching and instantiation involve aligning a schema from the schema library with events extracted by the schema extraction component, specifically for predicting supply chain disruptions. This process begins by instantiating one of the \(\mathbf{E}_{\text{schema}}\) from the integrated library or selecting the extracted event \(E_{\text{ext}}\) that best matches the schema event \(E_{\text{schema}}\). Subsequently, the task entails matching events in the \(\mathbf{E}_{\text{schema}}\) with their corresponding events in \(\mathbf{E}_{\text{ext}}\) extracted from the news dataset. Events in both \(\mathbf{E}_{\text{ext}}\) and \(\mathbf{E}_{\text{schema}}\) are organized in a highly structured manner, with parent events divided into child events. Events also contain temporal information, indicating that some events must precede others. Logical relationships are also defined: AND-gates connect all necessary child events for a parent event, OR-gates connect one or more needed child events, and XOR-gates indicate that only one child event can be present.

For example, a document about a raw material shortage in the EV battery supply chain might align with a "Supply Chain Disruption" schema in the schema library. Following the instantiation, a "notify suppliers" event in the schema might match with a graph \(G\) event describing a notification sent to cobalt suppliers. The "suppliers" participant of the schema event might match with the "cobalt suppliers" participant of the extracted event.

\subsection{Matching Process \& Techniques}

Our approach to event matching and instantiation involves several key steps and techniques to ensure accurate alignment between schema events and extracted events. This is particularly critical in the context of predicting supply chain disruptions, where precise event matching can provide actionable insights.

\subsubsection{Similarity Calculation}
To determine the similarity between schema events and extracted events, we calculate a similarity score based on semantic and structural similarities. Semantic similarity (\(\text{SemSim}\)) is computed using sentence transformers to encode the semantic content of events. Structural similarity (\(\text{StrSim}\)) takes into account the hierarchical and temporal relationships between events.

\textbf{Semantic Similarity:} We use a sentence transformer model to encode events into semantic vectors. The cosine similarity between these vectors provides a measure of how semantically similar two events are:
\begin{equation}
    \text{SemSim}(E_{\text{ext}}, E_{\text{schema}}) = \frac{\mathbf{v}_{\text{ext}} \cdot \mathbf{v}_{\text{schema}}}{\|\mathbf{v}_{\text{ext}}\| \|\mathbf{v}_{\text{schema}}\|}
\end{equation}
where \(\mathbf{v}_{\text{ext}}\) and \(\mathbf{v}_{\text{schema}}\) are BERT embeddings of extracted and schema events.

\textbf{Structural Similarity:} We consider the context of events within their respective hierarchies. For example, an event's predecessors and successors, its parent event, and its child events all contribute to its structural context. Events with similar structures in both schema and extracted graphs are more likely to match:
\begin{equation}
    \text{StrSim}(E_{\text{ext}}, E_{\text{schema}}) = \frac{|\mathbf{P}_{\text{ext}} \cap \mathbf{P}_{\text{schema}}|}{|\mathbf{P}_{\text{ext}} \cup \mathbf{P}_{\text{schema}}|}
\end{equation}
where \(\mathbf{P}_{\text{ext}}\) and \(\mathbf{P}_{\text{schema}}\) are the parameter sets for the extracted and schema events.

\subsubsection{Event Matching}
Once the similarity scores are calculated, we match each extracted event \(E_{\text{ext}}\) with the schema event \(E_{\text{schema}}\) that has the highest similarity score. This involves instantiating the schema event with information from the extracted event, ensuring that all relevant details and relationships are preserved.

\textbf{Example:} Consider a schema event "notify suppliers" in the context of a raw material shortage. An extracted event describing an email notification to cobalt suppliers would match this schema event if the similarity score is high. The instantiation process involves mapping the "suppliers" participant in the schema to the "cobalt suppliers" entity in the extracted event:
\begin{equation}
    \text{Instantiate}(E_{\text{matched}}, \mathbf{S}_{\text{schema}}) \rightarrow \mathbf{E}_{\text{inst}}
\end{equation}
where \(\mathbf{E}_{\text{inst}}\) is the instantiated event enriched with attributes from the schema.

\subsubsection{Consistency Checks}
After matching events, we perform consistency checks to ensure that the instantiated schema adheres to logical and temporal constraints. This includes verifying that:
\begin{itemize}
    \item All necessary child events are present (AND-gates).
    \item At least one required child event is present (OR-gates).
    \item Only one of the mutually exclusive child events is present (XOR-gates).
\end{itemize}
These checks ensure that the instantiated schema is logically coherent and temporally consistent:
\begin{equation}
    \text{ConsistencyCheck}(\mathbf{E}_{\text{inst}}, \mathbf{S}_{\text{schema}})
\end{equation}

\subsection{Continuous Improvement}
To enhance the accuracy and robustness of our matching and instantiation process, we incorporate continuous improvement through manual review and feedback from domain experts. This involves:
\begin{itemize}
    \item Validating the instantiated events with domain experts to ensure they accurately reflect real-world scenarios.
    \item Refining our models based on feedback, adjusting similarity metrics, and improving our semantic and structural encoding techniques.
    \item Iteratively updating our schema library and extraction models to incorporate new insights and improve performance.
\end{itemize}

By leveraging domain expertise and feedback, we continually refine our event matching and instantiation process, ensuring it remains effective and relevant for predicting and analyzing supply chain disruptions.

\begin{algorithm}[htbp]
\caption{Event Matching and Instantiation}
\label{alg:event_matching_instantiation}
\footnotesize
\begin{algorithmic}[1]
    \State \textbf{Input:} Extracted events $\mathbf{E}_{\text{ext}}$, schema library events $\mathbf{E}_{\text{schema}}$
    \State \textbf{Output:} Instantiated events $\mathbf{E}_{\text{inst}}$
    
    \State \textbf{Calculate Similarity} \Comment{Compute similarities}
    \For{each $E_{\text{ext}}$ in $\mathbf{E}_{\text{ext}}$}
        \For{each $E_{\text{schema}}$ in $\mathbf{E}_{\text{schema}}$}
            \State $\text{Sim}(E_{\text{ext}}, E_{\text{schema}}) \leftarrow \alpha \cdot \text{SemSim}(E_{\text{ext}}, E_{\text{schema}}) + \beta \cdot \text{StrSim}(E_{\text{ext}}, E_{\text{schema}})$
        \EndFor
    \EndFor

    \State \textbf{Match Events} \Comment{Match extracted events to schema events}
    \For{each $E_{\text{ext}}$ in $\mathbf{E}_{\text{ext}}$}
        \State $E_{\text{matched}} \leftarrow \underset{E_{\text{schema}}}{\operatorname{arg\,max}} \; \text{Sim}(E_{\text{ext}}, E_{\text{schema}})$
        \State $\mathbf{E}_{\text{inst}} \leftarrow \text{Instantiate}(E_{\text{matched}}, \mathbf{S}_{\text{schema}})$
        \State \text{Perform ConsistencyCheck}($\mathbf{E}_{\text{inst}}$, $\mathbf{S}_{\text{schema}}$)
    \EndFor

    \State \textbf{Continuous Improvement} \Comment{Manual review and feedback}
    \For{each $\mathbf{E}_{\text{inst}}$}
        \State $\text{UpdatedModels} \leftarrow \text{ValidateRefine}(\mathbf{E}_{\text{inst}})$
    \EndFor
    
    \State \textbf{Return:} Instantiated events $\mathbf{E}_{\text{inst}}$
\end{algorithmic}
\normalsize
\end{algorithm}

\section{Details of Disruption Prediction}
\label{app: disruption prediction}
Given an instantiated event graph $\mathbf{G}_{inst} = (N, E)$, where $N$ represents event nodes (e.g., specific supply chain activities) and $E$ denotes event-event temporal links (e.g., dependencies or sequences of activities), the goal is to classify whether unmatched schema events (nodes) could potentially occur within this graph.

Formally, let $I$ be the set of matched schema events within the graph. The task involves classifying each node in the remaining schema event nodes, represented by $N \setminus I$, as a missing event (positive or negative) given the instantiated graph.

To address the limitations of existing methods, we developed a novel approach that leverages the structural information within the schema graph and incorporates logic gates and hierarchies. Our approach consists of three stages:~(1)~schema-guided prediction,~(2)~constrained prediction, and (3)~argument coreference.

\subsection{Schema-Guided Prediction}
In this stage, we utilize a trained graph neural network specifically designed for schema graphs to score and select unmatched events in the instantiated graph. Key steps include:
\begin{itemize}
    \item \textbf{Graph Neural Network:} A GCN is trained on schema graphs to learn representations of nodes and edges. The propagation rule is given by:
    \begin{equation}
        \mathbf{H}^{(l+1)} = \sigma(\mathbf{A} \mathbf{H}^{(l)} \mathbf{W}^{(l)})
    \end{equation}
    where \(\mathbf{H}^{(l)}\) is the hidden state at layer \(l\), \(\mathbf{A}\) is the adjacency matrix, \(\mathbf{W}^{(l)}\) is the weight matrix, and \(\sigma\) is a non-linear activation function.
    \item \textbf{Node Scoring:} Using the learned representations, the GCN scores and selects unmatched events in the instantiated graph.
    \item \textbf{Prediction Output:} The first-stage prediction output consists of the most likely missing events.
\end{itemize}

\subsection{Constrained Prediction}
This stage applies logical constraints and hierarchical relations to refine the initial predictions from the schema-guided prediction stage. Key steps include:
\begin{itemize}
    \item \textbf{Logical Constraints:} We refine initial predictions (\(\hat{y}\)) to produce final predictions (\(\hat{y}'\)) that adhere to known rules:
    \begin{equation}
    \begin{aligned}
        \hat{y}' &= \underset{\hat{y}' \in \mathcal{Y}}{\operatorname{arg\,min}} \; \text{Constrain}(\hat{y}) \\
        & \text{subject to} \quad \mathcal{C}(\hat{y}') = \text{true}
    \end{aligned}
    \end{equation}
    where \(\mathcal{C}\) represents constraint sets. For example, a constraint might ensure that a major supplier's disruption increases risk for dependent manufacturers.
    \item \textbf{Hierarchical Relationships:} 
    \begin{itemize}
        \item \textbf{Child-to-Parent Propagation:} If a child event node is predicted or matched, its parent node is also predicted.
        \item \textbf{AND-Siblings Propagation:} If a predicted node has AND-sibling nodes, all its siblings are also predicted.
        \item \textbf{Iterative Refinement:} The constrained prediction approach is applied iteratively until no further nodes can be predicted.
    \end{itemize}
\end{itemize}

\subsection{Argument Coreference}
In this phase, we utilize coreference entity links and instantiated entities to generate predictions for the arguments associated with the predicted events. Key steps include:
\begin{itemize}
    \item \textbf{Coreference Links:} Coreference entity links specified in the schema are used to ensure consistency among entity mentions:
    \begin{equation}
    \begin{aligned}
    R_{ij} &= \underset{E_i, E_j \in \mathcal{E}}{\operatorname{arg,max}} \; \text{Coref}(E_i, E_j) \\
    & \text{subject to} \quad \text{Coref}(E_i, E_j) = \text{true}
    \end{aligned}
    \end{equation}
    where \((E_i, E_j)\) denotes each event pair and \(R_{ij}\) represents their relation.
    \item \textbf{Instantiated Entities:} Instantiated entities from the previous stages are leveraged to generate arguments for predicted events.
    \item \textbf{Final Output:} This stage produces the final prediction output, including both events and their arguments.
\end{itemize}

\section{Experiment Details}
\label{app: experiment details}
\subsection{Experiment Setup}
In Experiment \ref{exp: schema learning}, we evaluate the efficacy of three distinct Large Language Models (LLMs) in extracting hierarchical structures from our schema learning dataset. Leveraging domain expert knowledge, we annotate individual schemas for each article in our academic corpus using our proprietary system viewer and editor.~We then employ the methodology outlined in Appx.~\ref{app: schema merging} to synthesize these schemas into an integrated library. This combination of individual schemas and the integrated library serves as the ground truth for our hierarchical information extraction phase.
Our schema learning performance evaluation consists of two key components.~First, we compare the hierarchical information extracted by the three LLMs against our established ground truth. Second, we assess the consistency, accuracy, and completeness of the hierarchical structures derived from the textual content of each article in the schema learning dataset, with domain experts actively participating in this evaluation process.

In Experiment \ref{exp: disruption prediction performance}, we apply a similar annotation methodology to our news dataset as used for the schema learning dataset. However, annotating the news dataset presents unique challenges, as news reports typically do not explicitly elucidate the connections between events. Instead, they often employ speculative language to describe event interrelations. To ensure annotation accuracy, we heavily rely on domain knowledge derived from scenario documents throughout the annotation process. Subsequently, we utilize the ground truth extracted from these reports to evaluate our system's performance in predicting news report outcomes.

\subsection{Evaluation Metrics}
\textbf{Subjective~Schema~Learning}.~For subjective schema evaluation, we ensure that the event schemas generated from each paper and news report are consistent, accurate, and complete. The schema derived from academic papers demonstrates a logical hierarchical structure, while the schema produced from news reports presents a well-defined temporal sequence. Experts manually review the schemas to verify these attributes, providing qualitative feedback on the logical coherence and comprehensiveness of the extracted structures. Each schema is rated on a scale from 1 to 5, where 1 indicates poor quality and 5 indicates excellent quality.

\textbf{Objective~Disruption~Detection}.~We compare the instantiated schemas learned by our system with manually annotated ground truth to assess the degree of overlap. This comparison uses an evaluation metric similar to Smatch~\cite{cai2013smatch}, which involves breaking down both our schema and the ground truth into quadruples of the form \textit{relation(event1, event2, importance)}. For instance, the \textit{event} of \textit{Raw Material Mining} includes the \textit{subevent} of \textit{Lithium Mining} with an associated importance value, represented by the quadruple \textit{subevent(raw material mining, lithium mining, importance)}. Other relations include participants, gates, sequential events, etc.

To evaluate the results, we~1)~map the events in the learned schema $\mathbf{S}_{l}$ to those in the ground-truth schema $\mathbf{S}_{gt}$,~2)~establish a one-to-one mapping of quadruples between the learned schema $\mathbf{S}_{l}$ and the ground-truth schema $\mathbf{S}_{gt}$,~3)~calculate Precision, Recall, and F-score as follows:
\begin{equation}
\label{eq:eval_fscore}
    \text{Precision} = \frac{\text{number of matched quadruples in } \mathbf{S}_{l}}{\text{total quadruples in } \mathbf{S}_{l}}
\end{equation}

\begin{equation}
    \text{Recall} = \frac{\text{number of matched quadruples in } \mathbf{S}_{l}}{\text{total quadruples in } \mathbf{S}_{gt}}
\end{equation}

\begin{equation}
    \text{F-score} = 2\cdot \frac{\text{Precision}\cdot \text{Recall}}{\text{Precision} + \text{Recall}}
\end{equation}

\section{Disruption Prediction Case Studies}
\label{app: case studies}
\subsection{Case 1: Impact of the Inflation Reduction Act in August 2022}
In August 2022, the United States passed the Inflation Reduction Act, which included significant incentives for domestic EV battery production. This led to a rapid increase in investments but also highlighted potential material shortages, causing disruptions in the EV battery supply chain.

\vspace{-0.5em}
\textbf{System Prediction:}
Our system predicted the possibility of short-term material shortages by analyzing the market response data to the Inflation Reduction Act, monitoring global distribution reports of EV battery materials, and assessing the impact of increased domestic production incentives on the supply and demand balance.

\vspace{-0.5em}
\textbf{Outcome:}
The system identified the risks posed by the sudden increase in demand for battery materials, providing early warnings to stakeholders. This allowed them to take proactive measures such as securing long-term supply contracts and exploring alternative materials to mitigate potential shortages.

\subsection{Case 2: Lithium Supply Chain Disruption in Early 2023}
In early 2023, significant disruptions in the lithium supply chain were caused by escalating geopolitical tensions between Australia and China. As Australia is one of the world's largest suppliers of lithium, political factors heavily influenced its export policies, severely impacting the global supply chain for EV batteries, which rely heavily on lithium.

\vspace{-0.5em}
\textbf{System Prediction:} 
Our system accurately predicted the potential supply disruption by analyzing various news reports on geopolitical developments and export data. The system monitored news related to geopolitical tensions between Australia and China, analyzed export data indicating changes in Australia's lithium export policies, and integrated insights from scenario documents highlighting the dependence of the EV battery supply chain on Australian lithium exports.

\vspace{-0.5em}
\textbf{Outcome:}
The system flagged the risk of Australia's export restrictions to China, providing early warnings of potential disruptions in the EV battery supply chain.~This allowed stakeholders to proactively seek alternative sources and mitigate the impact on production.

\subsection{Case 3: Nickel and Cobalt Supply Issues in March 2023}
In March 2023, a major disruption in the global supply chain occurred due to large-scale worker strikes and regulatory changes in the Democratic Republic of Congo (DRC), a primary supplier of cobalt. Cobalt is crucial for EV batteries, and the disruption had a significant negative impact on the global supply chain.

\vspace{-0.5em}
\textbf{System Prediction:}
Our system successfully forecasted the potential supply chain interruptions by analyzing news reports on strike activities and updates on government regulations in the DRC. It also assessed historical data on cobalt supply and demand to identify vulnerabilities and integrated expert feedback on the impact of labor strikes and regulatory changes on cobalt production.

\vspace{-0.5em}
\textbf{Outcome:}
The system provided early warnings about the potential disruptions, enabling companies to adjust their supply chain strategies. This included diversifying sources of cobalt and increasing inventories to buffer against supply shortages.

\section{SHIELD's User Interface}
\label{app: Disruption Prediction User Interface Details}
The SHIELD user interface is designed to be intuitive and user-friendly, facilitating the efficient upload and analysis of news reports.

\textbf{News Report Upload}.~On the right side of the interface, users can upload their collected news report texts. It includes a text box for input and a submission button to upload the report (see Fig.~\ref{app:Fig_user_interface part1 user input}). Key features include:
\begin{itemize}
    \item \vspace{-0.5em} \textbf{Upload Box:} Allows users to paste or type their news report texts.
    \item \vspace{-0.5em} \textbf{Submit Button:} Initiates the analysis process once the report is uploaded.
    \item \vspace{-0.5em} \textbf{Uploaded Reports List:} Displays previously uploaded news reports, enabling users to review and compare past submissions easily.
\end{itemize}

\textbf{Disruption Analysis Results}.~After submitting a news report, users can view the real-time results of the disruption analysis on the left side of the interface (see Fig.~\ref{app:Fig_user_interface part2 visualization and edit}). The comprehensive overview of the analysis include:
\begin{itemize}
    \item \vspace{-0.5em} \textbf{Generated Schema:} Displays the hierarchical of events identified in the news report.
    \item \vspace{-0.5em} \textbf{Events List:} Lists all detected events and their details, allowing to see which events were identified and how they are connected.
    \item \vspace{-0.5em} \textbf{Evaluation Score:} Shows the real-time evaluation score, assessed against the schema library for accuracy and completeness.
    \item \vspace{-0.5em} \textbf{Schema Editing:} Allows to edit the generated schema. Users can make changes to the structure, relationships, and details of events.
    \item \vspace{-0.5em} \textbf{Regenerate Evaluation:} Users can choose to regenerate the evaluation score based on the edited schema, ensuring that the modifications are reflected in the updated score.
\end{itemize}

\section{Author Contributions}
\label{app:contribution}

\textbf{Schema Learning Dataset:}
\begin{itemize}
    \item Yuzhi Hu: Created the scenario document.
    \item Yifei Dong: Collected paper lists and Wikidata lists.
    \item Aike Shi: Collected Wikipedia lists and weekly report lists.
    \item Yifei Dong, Wei Liu, and Aike Shi: Extracted paragraphs from collected articles.
\end{itemize}

\textbf{Supply Chain News Dataset:}
\begin{itemize}
    \item Yuzhi Hu: Conducted data crawling and classification.
    \item Yuzhi Hu, Yifei Dong, Wei Liu, and Aike Shi: Labeled ground truth for the news dataset.
\end{itemize}

\textbf{Schema Learning:}
\begin{itemize}
    \item Yifei Dong and Aike Shi: Designed and modified prompts for generating structured information, designed the format for structured information, wrote scripts for converting structured information to SDF, and for generating structured information using Llama3.
    \item Yifei Dong, Aike Shi, Wei Liu, and Yuzhi Hu: Generated structured information using GPT-4o with zero-shot learning.
\end{itemize}

\textbf{Human Curation:}
\begin{itemize}
    \item Yifei Dong, Wei Liu, Aike Shi, and Yuzhi Hu: Curated LLM-generated SDFs.
    \item Yifei Dong and Aike Shi: Wrote scripts for schema merging.
\end{itemize}

\textbf{System Construction:}
\begin{itemize}
    \item Zhi-Qi Cheng: Provided guidance for system implementation, designed the system prototype, and performed system implementation.
    \item Yifei Dong and Aike Shi: Performed system debugging and testing.
\end{itemize}

\textbf{Evaluation:}
\begin{itemize}
    \item Wei Liu and Yifei Dong: Designed evaluation metrics.
    \item Aike Shi and Yifei Dong: Wrote evaluation scripts.
\end{itemize}

\vspace{2em}

\textbf{Frontend Interface:}
\begin{itemize}
    \item Aike Shi and Yifei Dong: Provided the design.
    \item Wei Liu: Optimized the design.
    \item Aike Shi: Implemented the interface.
\end{itemize}

\begin{figure*}[!ht]
  \centering
  \begin{subfigure}[b]{1\textwidth}
    \centering
    \includegraphics[width=\textwidth]{Figures/UI/ui_0.png}
    \caption{News report upload section of the user interface.}
    \label{app:Fig_user_interface part1 user input}
  \end{subfigure}
  \hfill
  \begin{subfigure}[b]{1\textwidth}
    \centering
    \includegraphics[width=\textwidth]{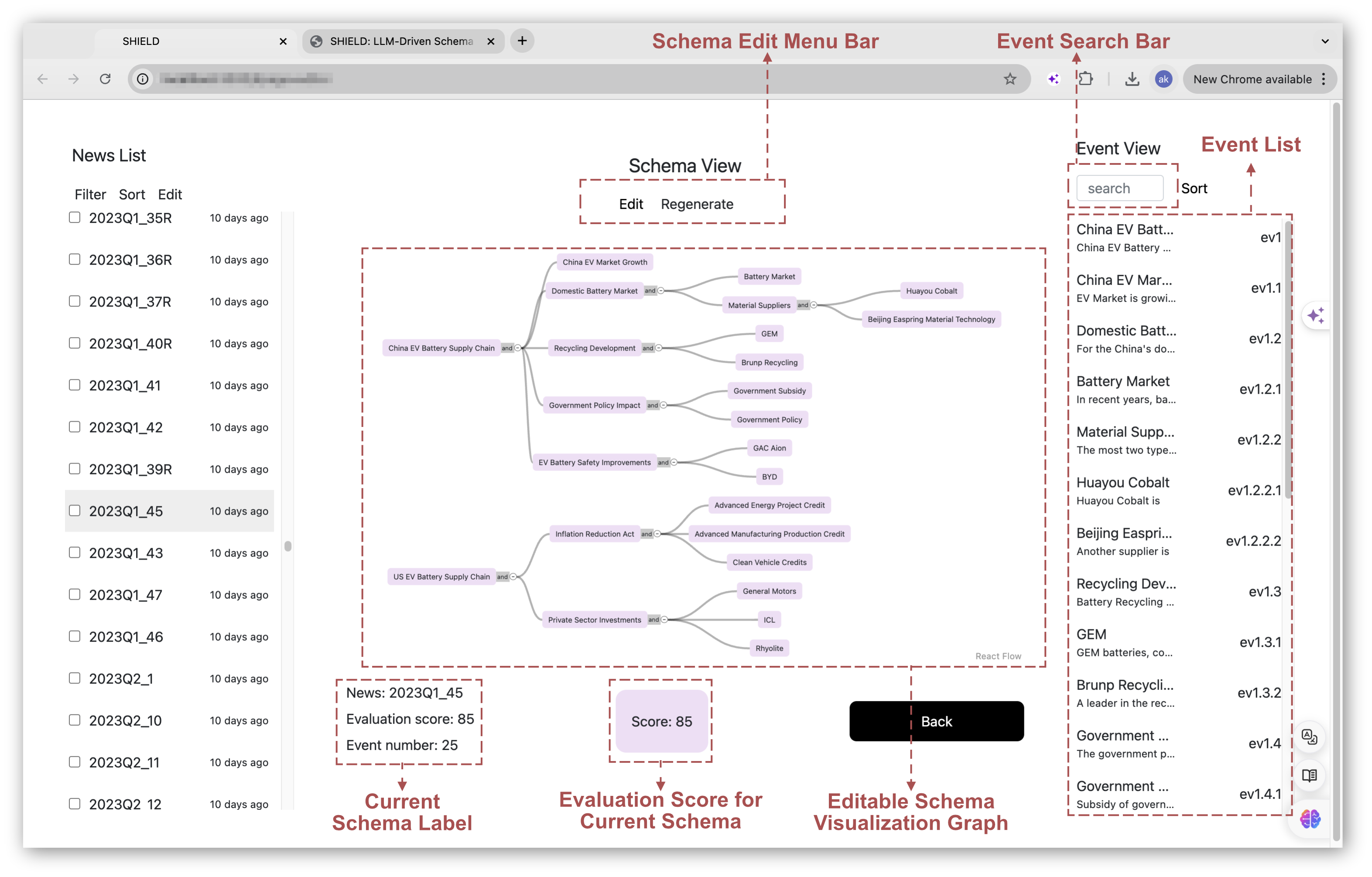}
    \caption{Visualization and editing of the final prediction results.}
    \label{app:Fig_user_interface part2 visualization and edit}
  \end{subfigure}
  \caption{User interface for the disruption prediction analysis in SHIELD.}
  \label{app:Fig_user_interface}
\end{figure*}

\textbf{Paper:}
\begin{itemize}
    \item Yifei Dong: Contributed to schema learning parts.
    \item Yuzhi Hu: Contributed to the news dataset and schema dataset parts and created and optimized Figs.~\ref{app:academic_dataset_example}, \ref{app:academic_dataset_example_wiki}, \ref{app:newsdataset_example}, and \ref{app:newsdataset_example_2}.
    \item Wei Liu, Yifei Dong, and Aike Shi: Contributed to Figs.~\ref{fig:motivation}, \ref{fig:schema_learning}, \ref{fig:prediction_pipeline}, \ref{fig:user_interface}, \ref{lst:example1}, \ref{app:fig:schema_management_system}, and \ref{app:Fig_user_interface}.
    \begin{itemize}
        \item Yifei Dong and Wei Liu designed all figures.
        \item Wei Liu created and optimized all figures.
        \item Aike Shi designed the user interface elements for Figs.~\ref{fig:user_interface}, \ref{app:fig:schema_management_system}, and \ref{app:Fig_user_interface}.
    \end{itemize}
    \item Jason O'Connor: Contributed to paper revisions and suggestions.
    \item Zhi-Qi Cheng: Organized and rewrote the entire paper, and supervised the modification of all figures.
\end{itemize}

\textbf{Presentation and Project Webpage:} \begin{itemize} \item Aike Shi and Zhi-Qi Cheng: Presented the paper. \item Yifei Dong: Built the project website. \item Aike Shi: Created the video demo. \item Wei Liu and Yuzhi Hu: Provided materials for the project website. \end{itemize}

\textbf{Feedback and Guidance:}
\begin{itemize}
    \item Jason O'Connor: Provided feedback and project guidance from a supply chain expert perspective.
    \item Zhi-Qi Cheng, Kate S. Whitefoot, and Alexander G. Hauptmann: Provided guidance and supervision for the entire project.
\end{itemize}

\end{document}